\documentclass[11pt]{article}

\usepackage[utf8]{inputenc}
\usepackage[T1]{fontenc}
\usepackage[a4paper,margin=1in]{geometry}
\usepackage{amsmath,amssymb,amsthm}
\usepackage{booktabs}
\usepackage{graphicx}
\usepackage{xcolor}
\usepackage{tikz}
\usepackage{pgfplots}
\pgfplotsset{compat=1.18}
\usetikzlibrary{arrows.meta,positioning,fit,shapes.geometric,calc,decorations.pathreplacing,patterns}
\usepackage{adjustbox}
\usepackage{tabularx}
\usepackage{caption}

\colorlet{dmfill}{blue!10}\colorlet{dmline}{blue!55!black}      % decision model
\colorlet{kgfill}{green!12}\colorlet{kgline}{green!40!black}    % knowledge graph
\colorlet{agfill}{orange!16}\colorlet{agline}{orange!75!black}  % agents
\colorlet{rtfill}{violet!7}\colorlet{rtline}{violet!65!black}   % ActiveGraph runtime
\usepackage{algorithm}
\usepackage{algpseudocode}
\usepackage[numbers,sort&compress]{natbib}
\usepackage[colorlinks=true,linkcolor=blue!50!black,citecolor=blue!50!black,urlcolor=blue!50!black]{hyperref}
\usepackage{microtype}
\usepackage{authblk}
\usepackage{orcidlink}

\newcommand{\chg}[1]{\,{\scriptsize(#1)}}
\newcommand{\model}{\textsc{Decision4Rec}}
\newcommand{\hb}{\,$\uparrow$}
\newcommand{\lb}{\,$\downarrow$}
\newcommand{\kgsys}{\model{}-KG}
\newcommand{\fullsys}{\model{}-Full}
\newcommand{\listsys}{\model{}-List}
\newcommand{\dm}{D}

\title{\model:\\ A Jev-Inspired Framework of Decision Models\\ on Knowledge and Active Graphs\\ for Multi-Agent Recommender Systems}

\author[1]{Marco Polignano\,\orcidlink{0000-0002-3939-0136}}
\author[2,3]{Het Darshan Mehta\,\orcidlink{0009-0000-8372-3877}}
\author[1]{Pasquale Lops\,\orcidlink{0000-0002-6866-9451}}
\author[2,3]{Ernesto~William~De~Luca\,\orcidlink{0000-0003-3621-4118}}
\author[1]{Giovanni Semeraro\,\orcidlink{0000-0001-6883-1853}}
\affil[1]{Università degli studi di Bari Aldo Moro, Piazza Umberto I, 70121 Bari, Italy}
\affil[2]{Otto-von-Guericke University, Universitätsplatz 2, 39106 Magdeburg, Germany}
\affil[3]{Leibniz Institute for Educational Media \textbar{} Georg Eckert Institute, Freisestr.~1, 38118 Braunschweig, Germany}
\affil[ ]{\small\texttt{marco.polignano@uniba.it}, \texttt{het.mehta@ovgu.de}, \texttt{pasquale.lops@uniba.it}, \texttt{giovanni.semeraro@uniba.it}, \texttt{deluca@ovgu.de}}

\date{}

\begin{document}
\maketitle

\begin{abstract}
Decision models, such as the commercial Jev, are language models that return one probability for each option of a question instead of generating text.% Used as released to rerank the candidates of a trained recommender, the open Kev-4B helps about as much as a random order. 
We present \model{}, an open and modular framework that makes decision models useful for recommendation. A knowledge graph writes the logs of each recommendation scenario as short statements on a defined scale, such as ``sold: top 10\%''. A pool of agents, proposes candidates, and a residual ranker keeps the best 30. A decision model, trained once on real choices of three scenarios, reads them through one interface of typed questions, and a fusion step combines its answers with the score of the ranker. Any model with this interface can be used, and we run the commercial Jev, the released open Kev and Kev models trained by us on recommendation requests.  The framework with a trained 4B decision model is 25 to 36\% above the best trained baseline in NDCG@10 in three sparse Amazon catalogues, one of them never used to train the decision model, and almost tie with baselines on the dense Steam dataset. Most of the gain comes from the pool, and with a trained recommender it comes from what the agents know about its candidates rather than from new candidates. The 27B decision model adds to the pool 5 to 6\% above the 0.8B model in three of four domains. A trained 0.8B open model beats the released Jev in eleven of fifteen settings and is never worse. On MovieLens-1M, a dense domain never seen in training, the gain is 5 to 7\%. Code, configurations and trained models are available at \url{https://github.com/marcopoli/decision4rec} and \url{https://huggingface.co/m-polignano}.
\end{abstract}

\section{Introduction}
\label{sec:intro}

Large language models (LLMs) are now part of many recommender systems. Most of them work as generators: they read the history of a user and a list of candidate items, and they write a ranked list, an explanation or a plan \citep{hou2024llmrank,wang2024macrec}. This brings four practical problems. The ranking changes when the same candidates come in another order \citep{hou2024llmrank,bito2026invarirank}. The output is text, so no probability says how sure the model is. Generation samples, so two runs can disagree. And the cost grows with the length of the text. A recommender does something simpler: it compares a small set of alternatives and estimates how likely each one is.

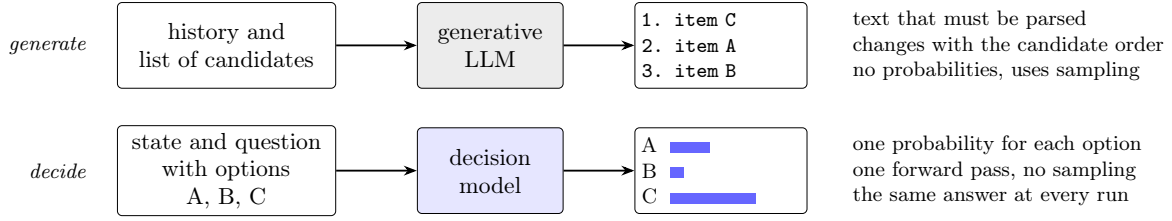
\begin{figure}[t]
\centering
\begin{adjustbox}{max width=\linewidth}
\begin{tikzpicture}[font=\small, x=1mm, y=1mm,
  box/.style={draw, rounded corners=2pt, align=center, minimum height=13mm, inner sep=3pt},
  note/.style={align=left, font=\footnotesize, text width=52mm},
  arr/.style={-{Stealth[length=2mm]}, thick}]
\node[box, text width=31mm] (in1) at (0,0) {history and\\list of candidates};
\node[box, text width=20mm, fill=gray!15] (llm) at (40,0) {generative\\LLM};
\node[box, text width=24mm, align=left, font=\footnotesize\ttfamily] (out1) at (75,0) {1. item C\\2. item A\\3. item B};
\node[note] (n1) at (121,0) {text that must be parsed\\changes with the candidate order\\no probabilities, uses sampling};
\draw[arr] (in1) -- (llm);
\draw[arr] (llm) -- (out1);
\node[box, text width=31mm] (in2) at (0,-19) {state and question\\with options A, B, C};
\node[box, text width=20mm, fill=blue!10] (dm) at (40,-19) {decision\\model};
\node[box, text width=24mm, align=left, font=\footnotesize] (out2) at (75,-19) {A\; \textcolor{blue!60}{\rule{6mm}{1.6mm}}\\B\; \textcolor{blue!60}{\rule{2mm}{1.6mm}}\\C\; \textcolor{blue!60}{\rule{13mm}{1.6mm}}};
\node[note] (n2) at (121,-19) {one probability for each option\\one forward pass, no sampling\\the same answer at every run};
\draw[arr] (in2) -- (dm);
\draw[arr] (dm) -- (out2);
\node[font=\footnotesize\itshape, anchor=east] at (-20,0) {generate};
\node[font=\footnotesize\itshape, anchor=east] at (-20,-19) {decide};
\end{tikzpicture}
\end{adjustbox}
\caption{Two ways to use a language model for ranking. A generative model writes a list as text. A decision model returns a probability for each option given by the caller.}
\label{fig:genvsdec}
\end{figure}

In September 2026 a different kind of model appeared. A \emph{decision model} does not generate text. It receives a state, for example the history of a user, and one or more questions with their options, and it returns one probability for each option in a single deterministic pass (Figure~\ref{fig:genvsdec}). The first one is the commercial Jev \citep{typesafe2026jev}, which we use in version \texttt{jev-1.13.0} through its API. Kev \citep{kev2026} offers open weights with the same interface. Early studies find that the choice probabilities of Jev are well calibrated \citep{deusser2026jevbench} and that its quality drops on fine rating scales \citep{gao2026ordinal}.

\paragraph{The problem.} The obvious use of a decision model in a recommender is to ask it which candidate the user will choose next. We tested this first (Section~\ref{sec:whytrain}). The released Kev-4B, reranking the 50 best items of a trained SASRec \citep{kang2018sasrec}, helps about the same users as a random order of the same items, and its fused score is not better than SASRec. Its request names the items but says nothing about the behavior of the scenario. With this behavior added as short statements, the same released model improves by 19 and 45\% in two of three domains, and training it on real choices improves it much more. Our first tests point to three needs: evidence about the scenario, good candidates, and training on real choices. Section~\ref{sec:gaps} shows seven gaps of this kind, each with its evidence and the requirement that it sets for a design.

\paragraph{Our idea.} A decision model returns $p(\text{option} \mid \text{state}, \text{options})$, a \emph{choice probability} in the sense of discrete choice theory \citep{luce1959choice,mcfadden1974conditional}. A choice depends on the person and on the alternatives. We write the history of the person in the state, and for each alternative we add what the scenario knows about it, as statements such as \emph{sold top 10\%; after last often; text top 50}. The words are positions on scales of the scenario, so that ``top 10\%'' can mean the same thing in a small shop and in a large one. We then train one decision model on real choices of several scenarios and reuse it in a new one.

\paragraph{The framework.} \model{} is a framework, not a single model (Section~\ref{sec:framework}). It has four modules with fixed contracts. A \emph{knowledge graph of the scenario} turns the logs into statements. A \emph{multi-agent pool} collects candidates from several points of view, with the scenario's own trained recommender among its agents, and records which agents found each candidate. A \emph{decision model} reads the best candidates with their text, statements and agents through one interface of typed questions. A \emph{fusion} step turns its answers and the candidate score into a ranking. Any model that speaks the interface fills the decision slot, commercial or open, released or trained, and changing it is one line of configuration. The modules are designed for the event-sourced runtime ActiveGraph \citep{nakajima2026log}, and the experiments of this paper run them as a batch pipeline.

\paragraph{Research questions.}
\begin{description}
\item[RQ1] In a domain whose records trained the decision model, does \model{} beat the strongest trained recommender of the domain?
\item[RQ2] In a domain never used to train the decision model, how close does \model{} come to a recommender trained there, with a catalogue only, with logs only, and with that recommender as one of its agents?
\item[RQ3] Which part brings the gain: the training, the statements, the candidates or the fusion with the trained score?
\item[RQ4] Can the probabilities be trusted, and what does one recommendation cost?
\end{description}

\paragraph{Contributions.}
\begin{enumerate}
\item An open, modular framework in which the decision model is an exchangeable part behind one interface of typed questions, released with code, configurations, trained models and the scripts that rebuild every dataset and training record.
\item Evidence that a released decision model adds to a trained recommender about as much as a random order, and that statements about the behavior of the scenario and training on real choices are what make it useful.
\item Statements on scales of the scenario, written by a knowledge graph that is never trained, and a decision model trained once on three scenarios and reused unchanged in a fourth.
\item A multi-agent pool that keeps the list of the trained recommender and lets a small residual ranker correct its order. Without a trained recommender, the candidates of the knowledge graph alone reach 81 to 92\% of a trained sequential model. With one, we show that the gain of the pool comes from what the agents know about its candidates, not from new candidates.
\item Three rules for training records of rerankers, each of which removed a real error from our first runs.
\item A comparison of the commercial Jev, the released Kev and our trained Kev models on identical requests, in five domains and against the best trained baseline of each. Training matters much more than price or size.
\end{enumerate}

\section{Related Work}
\label{sec:related}

\paragraph{Decision models.} Jev \citep{typesafe2026jev} answers typed questions instead of generating text: a choice among options, a position on a scale, or the probability that a statement is true. \citet{deusser2026jevbench} test it on 37 language understanding datasets and find well calibrated choice probabilities, while its yes or no probabilities are not reliable around a fixed threshold. \citet{gao2026ordinal} show that Jev and its open copies use fewer levels than the data when a rating scale becomes fine. Kev \citep{kev2026} has open weights and the same interface. AnyJev \citep{zhang2026anyjev} turns an open language model into a decision model without training, by reading the probabilities of the option labels. Its label readout allows at most 26 options. On a benchmark of general decisions, the accuracy of the Kev models grows with their size, from about 50\% at 0.8B to 77\% at 27B \citep{duan2026general}. These studies use decision models as released.
The closest work is \citet{lyu2026jevrerank}, who use Jev as a reranker of Amazon products with one choice question over the candidates. Their model is not trained on recommendation data, and they do not study open models, calibration, the order of the candidates or the source of the candidates. In a revised version with longer lists they conclude that a model made for decisions does not by itself recommend better. Our first test agrees with them (Section~\ref{sec:whytrain}), and this paper starts from that result.

\paragraph{Language models as recommenders.} Language models can rank candidates without training, but they are biased by the popularity of the items and by their position in the prompt \citep{hou2024llmrank}. P5 casts recommendation tasks as text generation \citep{geng2022p5}, TIGER generates the identifiers of the next item \citep{rajput2023tiger}, and LLaRA gives the language model the embeddings of a collaborative model \citep{liao2024llara}. These models are trained and used in one domain. LlamaRec \citep{yue2023llamarec} trains a language model to rerank the candidates of a sequential recommender and reads the scores of the option labels in one pass. A recent study finds that, with realistic candidates on Amazon data, trained rerankers and their fusions with collaborative filtering rarely beat collaborative filtering, and that a score written in the prompt is copied by the model \citep{wang2026recallceiling}. InvariRank changes a generative reranker so that its result does not depend on the order of the candidates \citep{bito2026invarirank}. A decision model gives a full probability vector for each question, so the effect of a position can be measured and corrected with a choice model instead of an architecture.

\paragraph{Statistics of the logs and transfer.} Sequential rules and item neighbors computed from the logs are strong recommenders without training \citep{ludewig2018evaluation}. STAR writes popularity and co-occurrence counts into the prompt of a language model \citep{lee2024star}. PrepRec moves to new domains with popularity over time and no item text \citep{wang2024preprec}, and RecGPT is trained on several domains and transfers through the history and the item text \citep{jiang2025recgpt}. None of them trains a model to read the statistics of a scenario together with the meaning of the items, and none gives choice probabilities.

\paragraph{Knowledge graphs and agents.} Knowledge-aware recommenders spread the preferences of a user over item attributes: RippleNet along the paths of the graph \citep{wang2018ripplenet}, KGAT with attention over its relations \citep{wang2019kgat}. Our knowledge graph is not trained. It writes counts and attributes as statements for a decision model. MACRec organizes recommendation as a collaboration of language agents \citep{wang2024macrec}, and RouteRec finds that aggregating the items of several recommender agents works better than choosing one agent per request \citep{zhou2026routerec}. In these systems the agents exchange generated text. Our agents return ranked lists, and a learned ranker aggregates them. \model{} is designed for an event-sourced agent runtime \citep{nakajima2026log}, in which a run can be replayed from its log (Section~\ref{sec:runtime}).

\paragraph{Choice models, calibration and contamination.} The choice axiom of Luce \citep{luce1959choice} and the conditional logit \citep{mcfadden1974conditional} link the choice probabilities over subsets to one scale of utility, and minorization-maximization estimates this scale from choices among many items \citep{hunter2004mm}. We apply these tools to a new respondent, a decision model. Temperature scaling \citep{guo2017calibration} and conformal sets \citep{sadinle2019lac} turn scores into usable confidence. Language models memorize public datasets such as MovieLens \citep{dipalma2025memorize}, which can make measured accuracy too high. Section~\ref{sec:risk} describes two controls for this.

\section{Background: Decision Models}
\label{sec:background}

A decision model $\dm$ receives a \emph{state} $s$, free text or structured data, and a set of named \emph{questions}. The answer to each question is a probability distribution, and no text is generated. A \texttt{choice} question has a set $O$ of named options, and the answer is a distribution $p_\dm(\cdot \mid s, q)$ over $O$. A \texttt{noul} question is a statement, and the answer is the probability that it is true. A \texttt{score} question has ordered levels, and the answer is a distribution over them. Figure~\ref{fig:request} shows a choice. This is the protocol of Jev \citep{typesafe2026jev}, and Kev uses the same one \citep{kev2026}. \model{} depends only on the protocol.

\begin{figure}[t]
\centering
\small
\fbox{\begin{minipage}{0.93\linewidth}
\textbf{State.} The user watched, in this order: Alien (1979), Blade Runner (1982), The Terminator (1984).\\[1mm]
\textbf{Question} (\texttt{choice}). Based on the user's interaction history, which candidate item is the user most likely to interact with next?\\[1mm]
\textbf{Options.} \texttt{A}: Aliens (1986) \quad \texttt{B}: Notting Hill (1999) \quad \texttt{C}: RoboCop (1987) \quad \texttt{none}: none of the listed candidates\\[1mm]
\textbf{Answer.} \texttt{A}: 0.52 \quad \texttt{B}: 0.03 \quad \texttt{C}: 0.38 \quad \texttt{none}: 0.07
\end{minipage}}
\caption{Example of a \texttt{choice} question and its answer. The probabilities are invented for illustration and are not measured.}
\label{fig:request}
\end{figure}

\paragraph{Three properties that we use.} (i) \emph{Determinism.} There is no sampling, so an answer is a function of the state and the question. On a GPU the open server can compute a question along two code paths whose results agree up to rounding. In our tests, identical requests gave identical answers. The hosted Jev does not (Section~\ref{sec:jev}). (ii) \emph{The caller defines the options}, so the system decides which alternatives are compared. (iii) \emph{Question isolation.} In the open implementation a question reads the state and itself, but not the other questions of the request \citep{kev2026}. The state is processed once, many questions are answered in one pass, and asking a question alone or together with two others changed its answer by at most 0.005 in our tests. Property (iii) makes it cheap to ask the same choice twice in different orders (Section~\ref{sec:ranking}).

\paragraph{What we do not assume.} The architecture and training of Jev are not public, and its calibration is reported as good for choices but not for yes or no questions or fine scales \citep{deusser2026jevbench,gao2026ordinal}. We therefore measure calibration on our own data and never compare probabilities across question types.

\begin{sloppypar}
\paragraph{Models.} We train Kev at 0.8B and 4B parameters with its own trainer: low-rank adapters (LoRA) and the choice head, started from the released weights (\texttt{jaredpalmer/kev-0.8b} and \texttt{jaredpalmer/kev-4b} on Hugging Face, built on \texttt{Qwen/Qwen3.5-0.8B-Base} and \texttt{Qwen/Qwen3.5-4B-Base}), on real next choices only. We also train \texttt{Kev-9B} and \texttt{Kev-27B} in the same way, on a random 8,000 of the records, to compare sizes on the same data. Kev-9B starts from \texttt{jaredpalmer/kev-9b} on \texttt{Qwen/Qwen3.5-9B-Base}, with the settings of the 4B model. Kev-27B starts from the adapter release of \texttt{jaredpalmer/kev-27b} (tag v1-lora) on \texttt{Qwen/Qwen3.8-27B}, with a learning rate of $2 \times 10^{-5}$, because its current release is a full fine-tune with no adapter to start from. \textbf{Jev} is used as released through its paid API. \textit{No output of a hosted model trains any model}.
\end{sloppypar}

\section{Gaps and Motivating Evidence}
\label{sec:gaps}

Before we describe \model{}, we show what is missing. Each gap comes with the evidence that it exists, measured with small diagnostic tests on validation and development users, and with the requirement (R1 to R7) that it sets for a design. %The tests are not part of \model{}. Section~\ref{sec:framework} names, for each module, the requirement that it meets.

\subsection{G1: a released decision model does not know the scenario}
\label{sec:whytrain}

\paragraph{The test.} We give the released Kev-4B the 50 best candidates of SASRec and ask one choice over them. For each user an oracle keeps the better of two rankings, SASRec or the decision model. The oracle is not a method: it only asks whether the decision model knows \emph{which} users it can help. As a null model, the same oracle uses a random order of the same candidates. The target is among the 50 candidates for 13.5 to 55\% of the users (Table~\ref{tab:whytrain}), so a perfect reranker could reach three to four times the NDCG@10 of SASRec. Yet the oracle with the decision model is only a little above the oracle with a random order (0.0802 against 0.0727 on Video Games), which already gives 64 to 88\% of the gain. The fused score is not significantly better than SASRec on any dataset.

\begin{table}[t]
\centering
\small
\caption{Does the released decision model know which users it can help? The released Kev-4B reranks the 50 best candidates of SASRec, 1,000 test users per dataset, NDCG@10. An oracle keeps, for each user, the better of two rankings; it is an upper bound for any switch between them. p: paired randomization test against SASRec. Last row: the part of the oracle gain over SASRec that a random order also gives.}
\label{tab:whytrain}
\begin{adjustbox}{max width=\linewidth}
\begin{tabular}{@{}lccc@{}}
\toprule
NDCG@10\hb{} & MovieLens-1M & Video Games & Musical Instr. \\
\midrule
SASRec alone & 0.1784 & 0.0593 & 0.0328 \\
users whose next item is among the 50 candidates\hb{} & 54.9\% & 23.0\% & 13.5\% \\
decision model fused with SASRec & 0.1806 (p = 0.41) & 0.0628 (p = 0.061) & 0.0317 (p = 0.31) \\
\addlinespace
oracle: the better of SASRec and the decision model & 0.2078 & 0.0802 & 0.0429 \\
oracle: the better of SASRec and a random order & 0.2044 & 0.0727 & 0.0408 \\
part of the oracle gain that a random order also gives\lb{} & 88\% & 64\% & 79\% \\
\bottomrule
\end{tabular}
\end{adjustbox}
\end{table}

\paragraph{Observations.} The request holds the titles and categories of the items, but nothing about how people behave in the scenario: which items sell now, or what people take after a given item. We added this behavior as the statements of Section~\ref{sec:evidence}. On the 30 best items of SASRecF, the released Kev-4B that reads them is better, read alone, than the same model without them by 19\% on Video Games and 45\% on Musical Instruments (p $\le$ 0.001), and not significantly different on Steam ($+1\%$, p = 0.78). After fusion it gains 13\% on Musical Instruments (p $<$ 0.001) and 0 to 2\% elsewhere.

\paragraph{Requirement R1.} The decision model must read the behavior of the scenario, and it must be trained on real choices to read it well.

\subsection{G2: a raw number means different things in different scenarios}
\label{sec:gap-scale}

\begin{figure}[t]
\centering
\begin{adjustbox}{max width=\linewidth}
\begin{tikzpicture}[font=\footnotesize, x=1mm, y=1mm, arr/.style={-{Stealth[length=2mm]}, thick}]
\draw[red!70!black, very thick, dashed] (62,-18) -- (62,5.5);
\node[font=\scriptsize\bfseries, text=red!70!black, anchor=south] at (62,5.5) {an item with 200 purchases};
\foreach \y/\name/\pa/\pb/\pc/\word in {0/{Lab supplies\\ (small scenario)}/18/52/96/{overall popularity: top 10\%}, -13/{Video games\\ (large scenario)}/45/115/132/{overall popularity: top half}} {
  \node[anchor=east, align=right] at (-2,\y) {\name};
  \fill[green!6] (0,\y-2.8) rectangle (\pa,\y+2.8);
  \fill[green!20] (\pa,\y-2.8) rectangle (\pb,\y+2.8);
  \fill[green!38] (\pb,\y-2.8) rectangle (\pc,\y+2.8);
  \fill[green!58] (\pc,\y-2.8) rectangle (145,\y+2.8);
  \draw[kgline] (0,\y-2.8) rectangle (145,\y+2.8);
  \foreach \x in {\pa,\pb,\pc} { \draw[kgline] (\x,\y-2.8) -- (\x,\y+2.8); }
  \node[font=\scriptsize] at ({\pa/2},\y) {rare};
  \node[font=\scriptsize] at ({(\pa+\pb)/2},\y) {top half};
  \node[font=\scriptsize] at ({(\pb+\pc)/2},\y) {top 10\%};
  \node[font=\scriptsize] at ({(\pc+145)/2},\y) {top 1\%};
  \node[circle, fill=red!70!black, draw=white, inner sep=1.8pt] at (62,\y) {};
  \node[draw, rounded corners=2pt, fill=white, anchor=west] (w) at (152,\y) {\word};
  \draw[arr] (146,\y) -- (w.west);
}
\node[anchor=west, font=\scriptsize] at (0,-20.5) {number of purchases of an item $\longrightarrow$};
\node[anchor=east, font=\scriptsize] at (145,-20.5) {borders: 50th, 90th and 99th percentile of each scenario};
\end{tikzpicture}
\end{adjustbox}
\caption{The same raw number means different things in different scenarios. Both bars use the same scale of purchases, and the red dot is the same item with 200 purchases. In a small scenario that sells lab supplies it is one of the best sellers (top 10\%); in a large video games scenario it is an ordinary item (top half). A statement gives the position of the item inside its scenario. The borders are invented for illustration.}
\label{fig:statements}
\end{figure}
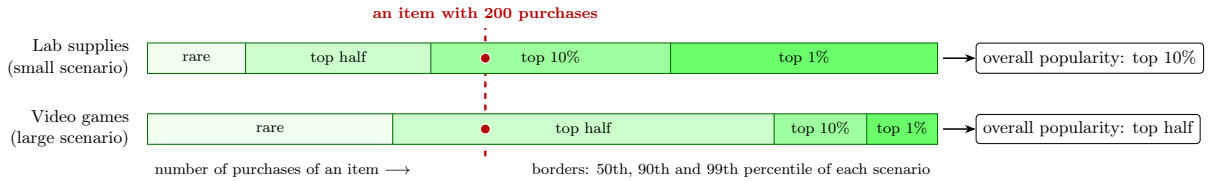

STAR \citep{lee2024star} gives behavior to a language model by writing counts into the prompt. A raw count, however, has no fixed meaning. Two hundred purchases make an item a best seller in a small shop of lab supplies and an ordinary item in a large shop of video games (Figure~\ref{fig:statements}). Our hypothesis is that a model trained on some scenarios reads the counts of a new scenario on the wrong scale, which would prevent reuse. We tested this with a control: a 0.8B decision model trained and read with raw counts instead of levels. In six settings it was never worse (read alone 0 to $+9\%$, fused $-1$ to $+4\%$), and the only significant difference favours raw counts (Industrial and Scientific, read alone, $+9\%$, p = 0.001). In our domains the scales are therefore a design choice that keeps the statements short and readable, not the reason why a decision model trained elsewhere works in a new scenario.

\paragraph{Requirement R2.} Behavior must reach the decision model in a form it can read in any scenario. We use positions on scales of the scenario; the control above shows that raw counts serve as well.

\subsection{G3: a reranker can only order the candidates it receives}
\label{sec:gap-recall}

A reranker reads a short list, here 30 items. If the item that the user takes next (the \emph{target}) is not in the list, no reranker can put it first. With the 30 best items of SASRecF, the trained recommender of our domains, the target is in the list for only 11.9 to 23.4\% of the validation users (Table~\ref{tab:mixes}). Other sources look at the user from other points of view: what usually follows the last purchase, what is bought with the history, what has a similar text. A scenario without a trained recommender has only such sources.

\begin{table}[t]
\centering
\small
\caption{Four simple ways to combine SASRecF with other sources of candidates, development validation users. First row of each block: SASRecF alone. Other rows: change against SASRecF on the same users. $^\ast$: the 95\% bootstrap interval of the difference excludes zero (computed for row (d) only). (a) the best of eight weights, chosen on the same users, so the row is optimistic. (c) weights tuned on one half of the users and reported on the other half. (d) trained on 15,000 other users of the domain.}
\label{tab:mixes}
\begin{adjustbox}{max width=\linewidth}
\begin{tabular}{@{}lcccc@{}}
\toprule
 & Video Games & Musical Instr. & Steam & Industrial and Sci. \\
\midrule
\multicolumn{5}{@{}l}{\emph{Users whose target is in the list of 30 (recall@30)}\hb}  \\
SASRecF alone, its 30 best items & 21.8\% & 13.0\% & 23.4\% & 11.9\% \\
(a) SASRecF score plus the score of the logs & $0\%$ & $-7\%$ & $-2\%$ & $-1\%$ \\
(b) rank fusion of all sources, equal weights & $-6\%$ & $-19\%$ & $-29\%$ & $-4\%$ \\
(c) rank fusion of all sources, tuned weights & $+2\%$ & $-2\%$ & $-11\%$ & $-2\%$ \\
\addlinespace
\multicolumn{5}{@{}l}{\emph{NDCG@10 over the whole catalogue}\hb} \\
SASRecF alone & 0.0632 & 0.0381 & 0.0657 & 0.0404 \\
(a) SASRecF score plus the score of the logs & $+2\%$ & $+8\%$ & $-4\%$ & $+4\%$ \\
(d) a ranker learned from scratch & $+3\%$ & $-16\%^\ast$ & $+1\%$ & $+20\%$ \\
\bottomrule
\end{tabular}
\end{adjustbox}
\end{table}

\paragraph{Evidence: simple mixes fail.} Before building the pool of Section~\ref{sec:pool}, we tried four simple designs (Table~\ref{tab:mixes}). (a) Adding the fused score of the logs (Section~\ref{sec:candidates}) to the score of SASRecF keeps fewer targets in every domain and loses on Steam. (b) Reciprocal rank fusion \citep{cormack2009rrf} of all lists with equal weights loses 4 to 29\% of the targets of SASRecF, because weak sources push out its items. (c) Tuned weights still lose in three of four domains. (d) A logistic ranker learned from scratch over all candidates is unstable, from $-16\%$ (significant) to $+20\%$. Every simple mix loses part of what SASRecF already finds, often more than it adds. Section~\ref{sec:pool} adds a later result: with a trained recommender, what the other sources contribute is information about its candidates, not new candidates.

\paragraph{Requirement R3.} Without a trained recommender, the candidates must come from the logs. With one, its list must be kept whole, and a ranker may only correct its order with what the other sources know about each candidate, so that it can always fall back to it.

\subsection{G4: training records for a reranker are easy to get wrong}
\label{sec:gap-records}

A cue that exists only in training gives a model that looks good on held-out records and fails on real users. Our first two exports of training records had such cues. The first one added the target to the candidates when the first stage had missed it, so the answer was the one item that the first stage did not like, an odd one out that a model can learn to pick. The second one took the third-to-last item as the target, but its candidates came from a recommender that had learned every earlier step of every history, so the first stage already knew the answer. A third cue is easy to add by mistake: statistics of the logs that count the target itself.

\paragraph{Requirement R4.} Training records must hold no cue that is absent at test time, and each rule that ensures this must be checked by a test in the code.

\subsection{G5 and G6: closed models and fixed pipelines}
\label{sec:gap-framework}

A result that depends on one closed model cannot be checked once the provider changes it, and the hosted Jev does not even answer the same request twice in the same way (Section~\ref{sec:jev}). Published systems also wire one model into one pipeline, so a team cannot compare or upgrade the model, reuse a trained model in a new scenario, or replace one part without rebuilding the rest.

\paragraph{Requirements R5 and R6.} Every result must be obtainable with open weights and recomputable from stored answers (R5). The system must be made of modules with fixed contracts and one interface for every decision model, so that a model is exchanged with one line of configuration and trained once for all scenarios (R6).

\subsection{G7: the content of the items is read only in part}
\label{sec:gap-content}

Descriptions, game tags and film plots are what a recommender can use for items with few interactions and in a new scenario, and they are what a language model reads best. In the version evaluated here, content enters through short texts: the text agent and the text statements encode the first 300 characters of each item, and the decision model reads the first 90, which keep the title and the category path. At 90 characters the description is reached for only 1 to 15\% of the items of the Amazon domains, and at 250 characters for 42 to 97\% (Additional Materials, AM2).

\paragraph{Requirement R7.} Content must reach both stages: the agents of the pool and the questions to the decision model. Here the requirement is met only in part. RQ3 tests a decision model that reads 250 characters per item.

\section{The \model{} Framework}
\label{sec:framework}

\subsection{Overview}

A \emph{recommendation scenario} is a platform with its catalogue, its users and their logs, for example a shop of video games or a service for lab supplies. Let $\mathcal{I}$ be the catalogue and $H_u = (i_1, \dots, i_T)$ the history of user $u$ in order of time. The task is to rank the items by the probability that each one is the next interaction $i_{T+1}$. Every item has a text, and the logs hold the histories of all users up to now.

\begin{figure}[t]
\centering
\begin{tikzpicture}[font=\footnotesize, x=1mm, y=1mm,
  mod/.style={draw, rounded corners=2pt, line width=0.8pt},
  ttl/.style={font=\footnotesize\bfseries, anchor=west},
  txt/.style={font=\scriptsize, align=center},
  num/.style={circle, fill=#1, text=white, font=\scriptsize\bfseries, inner sep=0.5pt, minimum size=3.6mm},
  chip/.style={draw=agline, fill=white, rounded corners=1pt, font=\scriptsize, text width=30.5mm, align=center, minimum height=3.8mm, inner sep=0.3mm},
  plug/.style={draw=dmline, fill=white, rounded corners=1pt, font=\scriptsize, text width=32mm, align=center, minimum height=3.8mm, inner sep=0.3mm},
  ctr/.style={font=\scriptsize\itshape, align=center, text=black!75},
  arr/.style={-{Stealth[length=2mm]}, thick}]
\draw[rtline, line width=0.9pt, rounded corners=3pt, fill=rtfill] (16.5,51) rectangle (157.5,-47.5);
\node[anchor=north west, text=rtline, font=\scriptsize\bfseries] at (18,50.2) {ActiveGraph runtime: every step is an event in the log};
\node[draw, rounded corners=2pt, fill=gray!10, txt, text width=15mm, inner sep=0.5mm, minimum height=10mm] (ev) at (7.3,18) {\textbf{Events}\\ new\\ interactions};
\node[draw, rounded corners=2pt, fill=gray!10, txt, text width=15mm, inner sep=0.5mm, minimum height=10mm] (cat) at (7.3,-2) {\textbf{Catalogue}\\ text and\\ attributes};
\draw[mod, draw=kgline, fill=kgfill] (18.5,35) rectangle (53,-26);
\node[num=kgline] at (21,31.8) {1}; \node[ttl] at (22.6,31.8) {Knowledge graph};
\begin{scope}[every node/.style={circle, draw=kgline, fill=white, inner sep=0pt, minimum size=3.6mm, font=\scriptsize}]
\node (i1) at (24,25) {$i_1$}; \node (i2) at (47.5,25) {$i_2$};
\node (i3) at (24,17) {$i_3$}; \node (i4) at (47.5,17) {$i_4$};
\node (i5) at (24,9.5) {$i_5$}; \node (i6) at (47.5,9.5) {$i_6$};
\end{scope}
\node[draw=kgline, fill=white, rounded corners=1pt, inner sep=0.5mm, font=\scriptsize] (e1) at (35.75,4.2) {brand: X};
\draw[-{Stealth[length=1.4mm]}, kgline, thick] (i1) -- node[above=-0.3mm, font=\tiny, text=black]{follow-up} (i2);
\draw[kgline, thick, dashed] (i3) -- node[above=-0.3mm, font=\tiny, text=black]{co-purchase} (i4);
\draw[kgline] (i5) -- (e1); \draw[kgline] (i6) -- (e1);
\node[draw=kgline, fill=white, rounded corners=1pt, font=\tiny, align=left, inner sep=0.8mm] at (35.75,-7.5) {overall popularity: top 5\%\\ follow-up: often\\ co-purchase: 2\\ text similarity: top 50};
\node[txt] at (35.75,-20) {statements on scales\\ of the scenario,\\ never trained};
\draw[mod, draw=agline, fill=agfill] (56,35) rectangle (90,-26);
\node[num=agline] at (59,31.8) {2}; \node[ttl] at (60.6,31.8) {Multi-agent pool};
\node[chip, line width=1.1pt] at (73,26) {trained recommender};
\node[chip] at (73,21.5) {follow-up};
\node[chip] at (73,17) {co-purchase};
\node[chip] at (73,12.5) {text similarity};
\node[chip] at (73,8) {recent popularity};
\node[chip] at (73,3.5) {rare attributes};
\node[chip] at (73,-1) {overall popularity};
\node[chip, dashed] at (73,-5.5) {perspectives (optional)};
\draw[agline, thick, -{Stealth[length=1.6mm]}] (73,-7.6) -- (73,-10.6);
\node[draw=agline, fill=white, rounded corners=1pt, txt, text width=28mm, inner sep=0.6mm, line width=0.8pt] (rr) at (73,-15) {\textbf{residual ranker}\\ corrects the trained\\ score, keeps 30};
\draw[mod, draw=dmline, fill=dmfill] (93,35) rectangle (128,-1);
\node[num=dmline] at (96,31.8) {3}; \node[ttl] at (97.6,31.8) {Decision model};
\node[txt, text width=31mm] at (110.5,16.5) {reads the history, item text, statements and ``found by''\\[1.5mm] one probability per candidate, one pass};
\node[font=\scriptsize\bfseries, text=dmline] at (110.5,3) {exchangeable slot};
\draw[dmline, line width=0.9pt] (110.5,-1) -- (110.5,-4.2);
\node[plug] at (110.5,-6.3) {Jev, API (released)};
\node[plug] at (110.5,-10.8) {Kev-4B (released)};
\node[plug, line width=1.1pt] at (110.5,-15.3) {Kev-0.8B, 4B (trained)};
\node[plug] at (110.5,-19.8) {Kev-27B (trained)};
\node[txt, text=dmline] at (110.5,-23.8) {one configuration line};
\draw[mod, draw=rtline, fill=white] (132,35) rectangle (156.5,-1);
\node[num=rtline] at (135,31.8) {4}; \node[ttl] at (136.6,31.8) {Fusion};
\node[txt, text width=22mm] at (144.25,16) {decision fused with the ranker score\\[1.5mm] ranking, with the evidence that was read};
\draw[arr] (ev.east) -- (18.5,18); \draw[arr] (cat.east) -- (18.5,-2);
\draw[arr] (53,5.5) -- (56,5.5);
\draw[arr] (90,14) -- (93,14);
\draw[arr] (128,14) -- (132,14);
\draw[arr, kgline] (35.75,35) -- (35.75,40.5) -- node[above, font=\scriptsize, text=black, pos=0.5]{statements of every candidate} (110.5,40.5) -- (110.5,35);
\draw[arr, agline, dashed] (rr.south) -- (73,-29.5) -- node[above, font=\scriptsize, text=black, pos=0.86]{ranker score} (144.25,-29.5) -- (144.25,-1);
\node[ctr] at (35.75,-36.5) {out: statements,\\ candidate scores};
\node[ctr] at (73,-36.5) {out: ranked lists,\\ ``found by'', top 30};
\node[ctr] at (110.5,-36.5) {in: state, typed questions;\\ out: probabilities};
\node[ctr] at (144.25,-36.5) {out: ranking\\ and its evidence};
\node[draw=rtline, fill=violet!20, rounded corners=1pt, font=\scriptsize, minimum width=137mm, minimum height=4.6mm] at (87.5,-43.5) {\textbf{Event log}: replay, counterfactuals, explanations by removing one statement};
\end{tikzpicture}
\caption{Architecture of \model{}: four modules with fixed contracts (in italics), designed for the ActiveGraph runtime and run as a batch pipeline in the experiments. (1) The knowledge graph of the scenario, never trained, writes statements on scales of the scenario. (2) The multi-agent pool, with the trained recommender as one agent; a residual ranker corrects its score and keeps 30 candidates. (3) The decision model, a slot that any model with the typed-question interface can fill. (4) Fusion with the ranker score.}
\label{fig:architecture}
\end{figure}
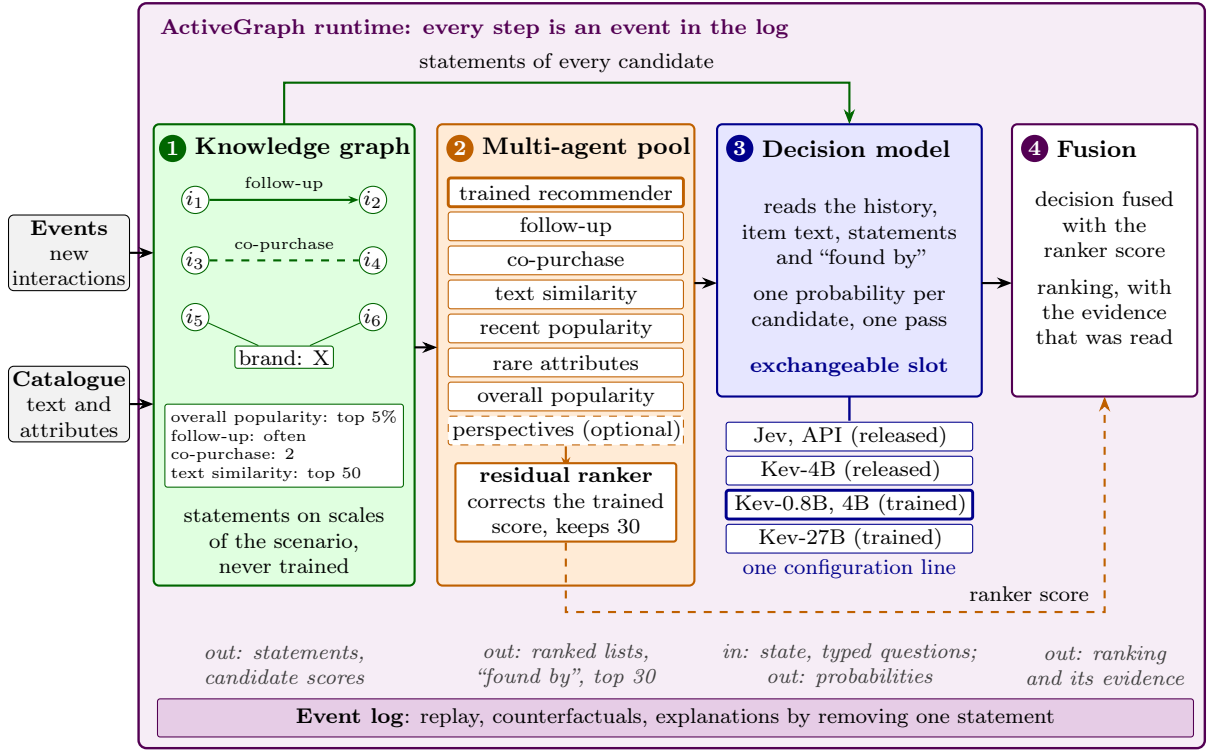

Figure~\ref{fig:architecture} shows the four modules. The knowledge graph of the scenario writes the logs as statements (Section~\ref{sec:evidence}). Agents, among them the trained sequential recommender of the scenario when there is one (the \emph{trained recommender}, SASRecF in our experiments), build a pool of candidates, from which a residual ranker keeps 30 (Section~\ref{sec:pool}). The decision model, trained once on real choices (Section~\ref{sec:training}), reads them, and a choice model turns its answers into a ranking that is fused with the score of the ranker (Section~\ref{sec:ranking}). Each module has a small contract: the knowledge graph returns statements and candidate scores, each agent returns a ranked list, the decision model returns one probability per option, and the fusion returns a ranking. A part that respects its contract can be replaced without changing the others (requirement R6). This is what lets the experiments measure each module alone, add or remove agents, and run Jev, the released Kev and the trained Kev on identical requests.

\subsection{The knowledge graph of the scenario, read as statements}
\label{sec:evidence}

This module meets requirement R2: behavior reaches the decision model as positions inside the scenario.

\paragraph{What the graph keeps.} From the training part of every history, the scenario keeps four things. The first is the purchases of each item with their times. The second is the directed transitions $a \rightarrow b$ within 10 steps of the same user, weighted by one over the gap as in sequential rules \citep{ludewig2018evaluation}, with the 200 strongest successors of each item. The third is the co-occurrences within 5 steps, with the 100 strongest neighbors of each item. The last is one text vector per item, from the first 300 characters of its title, category path and description (encoder \texttt{BAAI/bge-small-en-v1.5}). Nothing is trained, and every count can be updated when a new event arrives.

\paragraph{Statements.} For a user with history $H$ at time $t$ and a candidate $i$, the graph writes five families of statements:
\begin{itemize}
\item \emph{overall popularity}: the purchases of $i$, as \emph{top 1\%}, \emph{top 10\%}, \emph{top half} or \emph{rare} among the items that sold at least once, or \emph{none yet};
\item \emph{recent popularity}: the same for the purchases in the 30 days before $t$;
\item \emph{follow-up}: how often other users chose $i$ right after the last three items of $H$ (weights $1, \frac12, \frac14$ from the most recent), from \emph{very often} to \emph{never};
\item \emph{co-purchase}: how many of the last 10 items of $H$ were bought together with $i$ (0 to 5, or 5+);
\item \emph{text similarity}: the rank of $i$ in the catalogue by the cosine between its text vector and those of the last three items of $H$ (\emph{top 10}, \emph{top 50}, \emph{top 500} or \emph{far}).
\end{itemize}
The borders of the levels are the 50th, 90th and 99th percentiles of the positive values in the scenario, so each word is a position inside it (Figure~\ref{fig:statements}). Each family is the signal of a strong training-free recommender \citep{ludewig2018evaluation}: popularity, trend, sequential rules, item co-occurrence and content similarity. Together they say how much an item sells, whether it sells now, what it follows, what it goes with and what it resembles. In the request each family is a short keyword, so the statements of one option take about 20 tokens and 30 options fit in one question. The state adds one line on how much data the scenario has and a legend of the keywords (Appendix~\ref{sec:appendix}). New families need only a function that returns a number for a candidate.

\paragraph{No statistic contains the answer.} The statistics use only the part of each history before the target. For evaluation the last two items of every user are left out, as for the trained baselines, and for training records one more (Section~\ref{sec:training}).

\subsection{Candidates from the knowledge graph}
\label{sec:candidates}

Without a trained recommender, the knowledge graph also proposes the candidates (requirement R3). We call this variant \kgsys{}. Four families give a score to every item. They are the follow-up weight $a(i)$, the co-purchase score $c(i)$, which is the co-occurrence with the last 10 items normalized by the degrees of the two items, the text rank $\mathrm{rk}(i)$, and the position $\mathrm{pos}(i)$ of $i$ among the 60 best sellers of the last 30 days. The fused score is
\begin{equation}
g(i) = w_a \frac{a(i)}{\max_j a(j)} + w_c \frac{c(i)}{\max_j c(j)} + w_t \max\!\left(0, 1 - \frac{\log \mathrm{rk}(i)}{\log 500}\right) + w_l \left(1 - \frac{\mathrm{pos}(i)}{60}\right),
\label{eq:gen}
\end{equation}
where each term is scaled to at most 1 for the user, a missing term is zero, and the items of the history are removed. A scenario with a catalogue only has the text term alone. Five settings of the weights gave the same recall on the validation users of the source domains (mean recall at 30 within 0.0016, AM3). We kept the one with the largest text weight, $(w_a, w_c, w_t, w_l) = (1, 1, 0.6, 0.3)$, because text is what a new scenario has.

Table~\ref{tab:candidates} compares the fused score with item kNN and sequential rules \citep{ludewig2018evaluation} and with SASRecF, a SASRec \citep{kang2018sasrec} whose item representation adds embeddings of catalogue attributes. The fused score finds more targets than both training-free methods in all four domains, but 18 to 24\% fewer than SASRecF. As a ranking of the whole catalogue it reaches 81 to 92\% of the NDCG@10 of SASRecF, with no training.

\begin{table}[t]
\centering
\small
\caption{Candidates from the knowledge graph, 1,000 validation users of each domain (development results). Top: share of users whose next item is among the 30 candidates (recall@30). Bottom: NDCG@10 of the same score used as a ranking of the whole catalogue. Industrial and Scientific was not used to choose anything.}
\label{tab:candidates}
\begin{adjustbox}{max width=\linewidth}
\begin{tabular}{@{}lcccc@{}}
\toprule
 & Video Games & Musical Instr. & Steam & Industrial and Sci. \\
\midrule
\multicolumn{5}{@{}l}{\emph{Users whose next item is among the 30 candidates}\hb} \\
fused score, Eq.~\ref{eq:gen} (used in \model{}) & 17.8\% & 9.9\% & 18.5\% & 9.3\% \\
item kNN (training-free baseline) & 16.3\% & 8.8\% & 18.3\% & 7.6\% \\
sequential rules (training-free baseline) & 15.1\% & 8.6\% & 18.4\% & 7.7\% \\
SASRecF (trained in the domain) & 21.8\% & 13.0\% & 23.4\% & 11.9\% \\
\addlinespace
\multicolumn{5}{@{}l}{\emph{NDCG@10 over the whole catalogue}\hb} \\
fused score, Eq.~\ref{eq:gen} & 0.0567 & 0.0318 & 0.0532 & 0.0370 \\
SASRecF (trained in the domain) & 0.0632 & 0.0381 & 0.0657 & 0.0404 \\
\bottomrule
\end{tabular}
\end{adjustbox}
\end{table}

\subsection{A multi-agent pool, with the trained recommender as one agent}
\label{sec:pool}

With a trained recommender, \model{} uses all its modules, and we call this variant \fullsys{}. The pool meets requirement R3, since the list of SASRecF enters whole and the ranker only corrects its order.

\paragraph{Agents.} The trained recommender adds its 50 best items. Each other agent adds its 20 best. Four agents read one scored family each (follow-up, co-purchase, text similarity, recent popularity), and one reads the fused score of Eq.~\ref{eq:gen}. One more proposes items that share rare catalogue attributes with the last three items, and the last proposes the best sellers. These eight agents form the pool of the main experiments. Most of them are simple retrieval rules. We call them \emph{agents}, as in work on multi-agent recommenders, because each proposes its own list from one point of view. Except the trained recommender, no agent is trained. Optional \emph{perspective agents} (complement, continuation, replenishment, loyalty, exploration) were tested in several versions (Additional Materials, AM5).

\paragraph{Residual ranker.} The agents propose a median of 108 to 141 items per user, too many for one question. A small \emph{residual ranker} orders them and keeps 30. The score of candidate $i$ is
\begin{equation}
r(i) = z_{\mathrm{SASRecF}}(i) + \lambda\, f(\mathbf{x}_i),
\label{eq:resid}
\end{equation}
where $z_{\mathrm{SASRecF}}(i)$ is the score of the trained recommender standardized over the candidates of the user, $\mathbf{x}_i$ says which agents found $i$ and at which rank, together with the numbers behind its statements, and $f$ is a linear correction that starts at zero. With $\lambda = 0$ the ranker returns the order of the trained recommender. The correction is trained with a softmax loss on 15,000 users of the scenario that are never evaluated, with their second-to-last item as target, and $\lambda$ and the number of steps are chosen on a fifth of these users.

\paragraph{What the decision model reads.} The decision model reads the 30 kept candidates with their text, their statements and one more line that names the agents that found them, for example ``found by: SASRecF \#5, follow-up''. It never reads the score of the trained recommender, which a language model tends to copy \citep{wang2026recallceiling}. Its records come from the three source domains only, so in a fourth domain it reads a pool and a scenario that it has never seen.

\begin{table}[t]
\centering
\small
\caption{The multi-agent pool, 1,000 validation users per domain (development results). Top: share of users whose next item is among the candidates; the second row is the ceiling for any ranker of the pool, the third row is the list that the decision model reads. Bottom: NDCG@10 over the whole catalogue, and the change of the residual ranker (Eq.~\ref{eq:resid}) against SASRecF; $^\ast$: the 95\% bootstrap interval of the difference excludes zero.}
\label{tab:pool}
\begin{adjustbox}{max width=\linewidth}
\begin{tabular}{@{}lcccc@{}}
\toprule
 & Video Games & Musical Instr. & Steam & Industrial and Sci. \\
\midrule
\multicolumn{5}{@{}l}{\emph{Users whose next item is among the candidates}\hb} \\
SASRecF alone, its 30 best items & 21.8\% & 13.0\% & 23.4\% & 11.9\% \\
pool, all candidates of all agents (median 108 to 141 items) & 33.6\% & 22.4\% & 37.1\% & 19.2\% \\
pool, the 30 items kept by the residual ranker & 22.5\% & 13.6\% & 24.5\% & 13.9\% \\
\addlinespace
\multicolumn{5}{@{}l}{\emph{NDCG@10 over the whole catalogue}\hb} \\
SASRecF alone & 0.0632 & 0.0381 & 0.0657 & 0.0404 \\
pool ordered by the residual ranker & 0.0686 & 0.0403 & 0.0695 & 0.0492 \\
change against SASRecF\hb{} & $+9\%$ & $+6\%$ & $+6\%$ & $+22\%^\ast$ \\
\bottomrule
\end{tabular}
\end{adjustbox}
\end{table}

\paragraph{Do the other agents find new targets?} Not more than SASRecF itself does with a longer list. All candidates of the pool contain the target for 54 to 72\% more validation users than the 30 best items of SASRecF (Table~\ref{tab:pool}). But the best items of SASRecF in the same number as the pool contain it for more users still: 38.1 against 33.6\% on Video Games, 25.6 against 22.4\% on Musical Instruments, 46.1 against 37.1\% on Steam and 22.1 against 19.2\% on Industrial and Scientific (p $\le$ 0.005). The 30 items kept by the ranker contain the target for 3 to 17\% more users than those of SASRecF, and their order is 6 to 22\% better in NDCG@10. Unlike the simple mixes of Section~\ref{sec:gap-recall}, the ranker never falls below SASRecF.

\paragraph{Candidates or information?} We trained the same ranker, with the same features and users, to order as many of the best items of SASRecF as the pool has, instead of the lists of the agents. On the test users it reaches the same NDCG@10 as the pool ($-2.8$ to $+2.4\%$, p $\ge$ 0.078) and keeps the target in its 30 for the same share of users (within 0.6 points). With a trained recommender, the gain of the pool therefore comes from what the agents know about each candidate, not from the candidates they add. Without a trained recommender, the candidates of the agents are the only ones.

\subsection{Training on real choices}
\label{sec:training}

This module meets requirements R1 and R4. The decision model of \kgsys{} reads the candidates of the knowledge graph, and that of \fullsys{} reads the pool. Both are trained in the same way.

\begin{figure}[t]
\centering
\begin{adjustbox}{max width=\linewidth}
\begin{tikzpicture}[font=\footnotesize, x=1mm, y=1mm,
  c/.style={draw=black!60, minimum width=11mm, minimum height=6mm, inner sep=0pt, font=\scriptsize},
  seen/.style={c, fill=black!22}, hist/.style={c, fill=black!7}, none/.style={c, draw=black!25, text=black!35},
  ckpt/.style={c, pattern=north east lines, pattern color=black!35},
  tgt/.style={c, line width=1.1pt, font=\scriptsize\bfseries}]
\foreach \x/\lab in {0/{$i_1$}, 13/{$i_2$}, 26/{\dots}, 39/{$i_{T-3}$}, 52/{$i_{T-2}$}, 65/{$i_{T-1}$}, 78/{$i_T$}} { \node[font=\scriptsize] at (\x,7) {\lab}; }
\node[anchor=east, align=right] at (-8,0) {trained recommender\\ (SASRecF)};
\foreach \x in {0,13,26,39,52} { \node[seen] at (\x,0) {}; }
\node[ckpt] at (65,0) {}; \node[none] at (78,0) {--};
\node[anchor=west, font=\scriptsize, align=left] at (86,0) {learns every step up to $i_{T-2}$;\\ $i_{T-1}$ only chooses its checkpoint};
\node[anchor=east, align=right, text=kgline] at (-8,-10) {\kgsys{},\\ training record};
\foreach \x in {0,13,26,39} { \node[seen] at (\x,-10) {}; }
\node[tgt, fill=kgfill, draw=kgline] at (52,-10) {target}; \node[none] at (65,-10) {--}; \node[none] at (78,-10) {--};
\node[anchor=west, font=\scriptsize, align=left, text=kgline] at (86,-10) {target $i_{T-2}$; the statistics leave out\\ the last three items of every user};
\node[anchor=east, align=right, text=dmline] at (-8,-20) {\fullsys{},\\ training record};
\foreach \x in {0,13,26,39,52} { \node[seen] at (\x,-20) {}; }
\node[tgt, fill=dmfill, draw=dmline] at (65,-20) {target}; \node[none] at (78,-20) {--};
\node[anchor=west, font=\scriptsize, align=left, text=dmline] at (86,-20) {target $i_{T-1}$, which SASRecF\\ never learned as a target (rule 3)};
\node[anchor=east, align=right] at (-8,-30) {test case\\ (every system)};
\foreach \x in {0,13,26,39,52} { \node[seen] at (\x,-30) {}; }
\node[hist] at (65,-30) {}; \node[tgt, fill=red!20, draw=red!60!black] at (78,-30) {target};
\node[anchor=west, font=\scriptsize, align=left] at (86,-30) {target $i_T$; the statistics leave out\\ the last two items of every user};
\node[seen, minimum width=5mm, minimum height=3.5mm] (l1) at (2,-40) {}; \node[anchor=west, font=\scriptsize] at (5.5,-40) {read: history and statistics};
\node[hist, minimum width=5mm, minimum height=3.5mm] at (46,-40) {}; \node[anchor=west, font=\scriptsize] at (49.5,-40) {history only};
\node[ckpt, minimum width=5mm, minimum height=3.5mm] at (72,-40) {}; \node[anchor=west, font=\scriptsize] at (75.5,-40) {checkpoint only};
\node[none, minimum width=5mm, minimum height=3.5mm] at (102,-40) {}; \node[anchor=west, font=\scriptsize] at (105.5,-40) {never seen};
\end{tikzpicture}
\end{adjustbox}
\caption{What each part of the system may see, for one user with history $i_1, \dots, i_T$. Each row is one use of the history; the colored cell is the target. Rule 1: the statistics of the knowledge graph stop before the target, for every user. Rule 2 (not drawn): a record is kept only if its target is among the candidates, and a missing target is never added. Rule 3: when the candidates come from a trained model, the target is an item that this model never learned as a target, so it cannot have memorized the answer. Users of every evaluation case are left out of \fullsys{}'s records.}
\label{fig:records}
\end{figure}
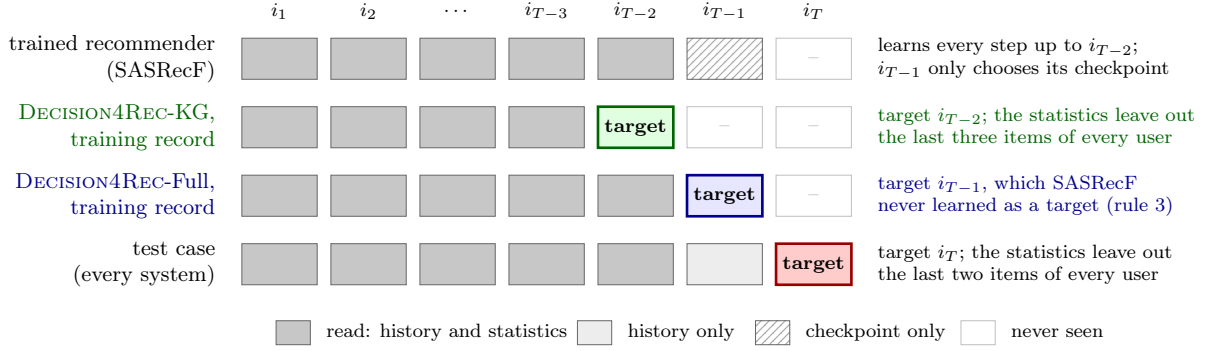

\paragraph{Records.} One record is one user (Figure~\ref{fig:records}). The state holds the last 10 items of the history, a line on the amount of data in the scenario, and the legend of the statements. The question is one choice over $K = 30$ candidates in random order, each with its text cut to 90 characters (enough for the title and the category path) and its statements. The label is the item that the user really chose next. We use Kev's trainer \citep{kev2026}: low-rank adapters and the choice head, from the released weights, one epoch, 8 records per step, with up to 15,000 users of each source domain and 500 more kept apart. For \kgsys{}, the statistics of a record come from all users of the scenario (40\% of the records) or from 20, 5, 1 or 0\% of them (15\% each), as the state line says, so that one model serves mature and new scenarios. Each family of statements is left out with probability 0.25.

\paragraph{Three rules.} Each rule removes one error of Section~\ref{sec:gap-records}, and each is checked by a test in the code. (1) \emph{No statistic contains the target}: the statistics of a record leave out, for every user, the target position and everything after it. (2) \emph{No target is added to the candidates}: records whose target is missing from the candidates are dropped, never completed. (3) \emph{No first stage was trained on the target}: the trained recommender learns from all but the last two items of every user, so the target of a \fullsys{} record is the second-to-last item, which the recommender used only to choose its checkpoint. Users of every evaluation set, and users that trained the ranker, are left out.

\subsection{Ranking as choice, and fusion}
\label{sec:ranking}

\paragraph{Two orders, one request.} One choice shows each candidate in one position, and the answer may depend on the position. We therefore ask the same choice twice, each time in a new random order, in one request (property iii of Section~\ref{sec:background}). Training records use one order.

\paragraph{Luce model.} The choice axiom of Luce \citep{luce1959choice} gives each item $i$ a positive \emph{worth} $v_i$: the probability of choosing $i$ from a set is its worth divided by the sum of the worths in the set. With worths 2, 1 and 1, the first item is chosen with probability 0.5. If the third item is removed, it is chosen with probability $2/3$, still twice as often as the second. Because each item keeps its worth in every set, a few choices give each item one score, and sorting by worth gives a ranking, as in the Plackett--Luce model \citep{plackett1975analysis}. A model may also prefer a position, for example the first one, so each position $k$ of a question $B$ gets a multiplier $b_k > 0$, with $b_1 = 1$. If $i_k$ is the item at position $k$ of $B$, the model gives position $k$ the probability
\begin{equation}
\pi_B(k) = \frac{v_{i_k}\, b_k}{\sum_{j=1}^{30} v_{i_j}\, b_j}.
\label{eq:luce}
\end{equation}
We write $\theta_i = \log v_i$ and choose $\theta$ so that $\pi_B$ is as close as possible to the answers $p_B$ of the decision model, minimizing the Kullback--Leibler divergence $\sum_B \sum_k p_B(k) \log \frac{p_B(k)}{\pi_B(k)}$ over the questions of the user. This is the likelihood of a multinomial logit with soft labels, a convex problem that minorization-maximization updates solve \citep{hunter2004mm} (Appendix~\ref{app:luce}). The benchmarked systems fit $b_k = 1$, since the two random orders already show each item in two positions. RQ4 measures the position effects.

\paragraph{Fusion.} The final score fuses the decision model with the residual ranker,
\begin{equation}
\mathrm{score}_u(i) = z(\theta_i) + \gamma\, z(r_u(i)),
\label{eq:fuse}
\end{equation}
where $z(\cdot)$ standardizes inside the 30 candidates and $\gamma$ is chosen on validation users.

\subsection{Risk control and contamination checks}
\label{sec:risk}

\paragraph{Risk control.} A confident wrong recommendation is worse than none. The probabilities $q_u = \mathrm{softmax}(\theta / \tau)$, with a temperature $\tau$ fitted on validation users \citep{guo2017calibration}, let the system abstain when it is unsure, or return a conformal set that should contain the target with a chosen probability \citep{sadinle2019lac}. The guarantee assumes exchangeable users, so RQ4 reports the measured coverage.

\paragraph{Contamination checks.}\label{sec:contamination} Public datasets such as MovieLens are memorized by language models \citep{dipalma2025memorize}. Two controls bound how much of a result memory explains: the same question with an empty state (a history-free prior) and with the item titles removed. Their results are in Section~\ref{sec:discussion}.

\subsection{Explicitly Active Graph Environment}
\label{sec:activegraph}

This module meets requirement R5 (Section~\ref{sec:gap-framework}): every step is recorded in a form that can be read and executed again.

\begin{figure}[t]
\centering
\begin{adjustbox}{max width=\linewidth}
\begin{tikzpicture}[font=\footnotesize, x=1mm, y=1mm,
  ev/.style={draw=rtline, fill=rtfill, rounded corners=2pt, text width=29mm, minimum height=9mm, align=center, inner sep=1mm, font=\scriptsize\ttfamily},
  cf/.style={draw, dashed, rounded corners=2pt, text width=29mm, minimum height=9mm, align=center, inner sep=1mm, font=\scriptsize},
  lab/.style={font=\scriptsize\itshape, text width=30mm, align=center},
  arr/.style={-{Stealth[length=2mm]}, thick}]
\node[ev] (a) at (0,0) {interaction.added};
\node[ev] (b) at (38,0) {logs.updated};
\node[ev] (c) at (76,0) {candidates.proposed};
\node[ev] (d) at (114,0) {ranking.estimated};
\node[ev] (e) at (152,0) {slate.emitted};
\foreach \u/\v in {a/b,b/c,c/d,d/e} { \draw[arr] (\u) -- (\v); }
\node[lab] at (19,10) {update counts, rules\\ and co-occurrence};
\node[lab] at (57,10) {agents propose,\\ pool keeps 30};
\node[lab] at (95,10) {decision, Luce fit\\ and fusion};
\node[lab] at (133,10) {calibration\\ and risk};
\node[cf] (f) at (38,-20) {fork: the same log\\ without interaction $k$};
\node[cf] (g) at (95,-20) {replay of the\\ same behaviors};
\node[cf] (h) at (152,-20) {counterfactual list};
\draw[arr, dashed] (a.south) |- (f.west);
\draw[arr, dashed] (f) -- (g);
\draw[arr, dashed] (g) -- (h);
\draw[{Stealth[length=2mm]}-{Stealth[length=2mm]}, thick] (e) -- node[right, font=\scriptsize, align=left]{effect of\\ interaction $k$} (h);
\end{tikzpicture}
\end{adjustbox}
\caption{The runtime. The boxes in the first row are the main events in the log, and the labels above the arrows say what the behaviors do between two events. The dashed row is a counterfactual: the log is forked, one interaction is removed, and the same behaviors run again; the difference between the two lists is the effect of that interaction.}
\label{fig:runtime}
\end{figure}
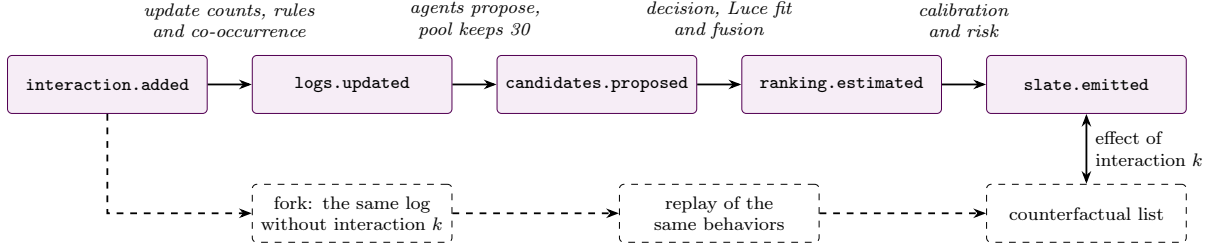

\paragraph{Method.} In the event-sourced runtime ActiveGraph \citep{nakajima2026log}, the source of truth is an event log that only grows; the working graph is computed from it, and small functions called \emph{behaviors} react to events and produce new ones (Figure~\ref{fig:runtime}). Each purchase updates the counts, transitions and co-occurrences of the items it touches, so the statements stay current and nothing is retrained. Each step (scenario, candidates, pool, decision, ranking, risk) is a separate behavior that can be changed alone. Because an open decision model returns the same probabilities for the same request, replaying the log gives the same recommendation; a hosted model such as Jev does not guarantee this (Section~\ref{sec:jev}). Asking again without one family of statements measures what each family changed (Appendix~\ref{app:explain}); forking the log without one interaction is possible in the runtime but is not measured in this paper.

\paragraph{Status in this paper.} The runtime is the deployment form of \model{}. In the released code it runs an earlier version of the agent (one request per user, an outside option and an expansion round); the experiments of Sections~\ref{sec:results} and~\ref{sec:jev} run the modules of this section as a batch pipeline with the same contracts, and their replay comes from the stored answers of the decision models, not from the event log. Porting Decision4Rec-Full to the runtime, with a check that it gives the same rankings as the batch pipeline on replayed users, is part of our ongoing work (Section~\ref{sec:ongoing}; the events are described in the Additional Materials, AM1).

\subsection{Runtime and cost}
\label{sec:runtime}

For one user the decision model receives one request with two questions of 30 options and generates no text. The knowledge graph needs no model call. Training is done once, on the source scenarios: about 0.3 seconds per record for the 0.8B model and 0.8 seconds for the 4B model on one RTX 3090, and about 4.2 seconds for the 9B model and 12.2 seconds for the 27B model on one NVIDIA DGX Spark (GB10, 128\,GB of unified memory), that is about 9 and 27 hours for their 8,000 records.

\section{Experimental Setup}
\label{sec:setup}

\subsection{Domains and their roles}

We use five datasets with item text (Table~\ref{tab:datasets}). Video Games, Musical Instruments and Steam are the \emph{source} domains: all training records of the decision model come from them. Industrial and Scientific and MovieLens-1M are \emph{held out}: no record, weight or design choice comes from them, except in two variants that say so (Sections~\ref{sec:systems} and~\ref{sec:rq2}). The Amazon domains come from Amazon Reviews 2023 \citep{hou2024blair}, Steam from \citet{kang2018sasrec} and MovieLens-1M from \citet{harper2015movielens}.

\paragraph{Selection of the sources.} A decision model that is trained once and reused should meet more than one platform and more than one density of data. Video Games and Musical Instruments are sparse Amazon catalogues (21 to 32 interactions per item) with long product texts, the typical case of a shop. Steam is another platform with a dense catalogue (134 interactions per item). Video Games and Steam both sell games, so the same words of the statements must be read on two very different scales. We stop at three sources because each needs its own trained recommender, knowledge graph and 15,000 training records on one GPU.

\paragraph{Motivation for held-out domains.} They test reuse at two distances. \emph{Industrial and Scientific} is near in form and far in content: the same platform and the same kind of text, but supplies bought for work, and the sparsest catalogue (16 interactions per item). It is the case of a shop that opens a new category. \emph{MovieLens-1M} is far in every respect: films instead of products, ratings instead of purchases, and long histories (165 interactions per user against 8 to 12). It is the hardest test of reuse here, it shows where reuse stops, and it is a standard dataset of the field. Its results are reported apart and do not enter the main claims. Because MovieLens-1M has only titles and genres, we also build an \emph{enriched} version with plots from \citet{spillo2026bingewatch} and Wikidata properties \citep{wikidata} (director, cast, writer, studio, series, previous and next film, and others), each found for 98\% of the films. Users, sequences and evaluation cases do not change.

\begin{table}[t]
\centering
\small
\caption{Datasets after preprocessing (5-core, ordered by time). For Steam we keep a random sample of 100,000 users of the 5-core. The last two columns give the mean number of interactions per user and per item: a catalogue is \emph{sparse} when its items have few interactions each (the Amazon domains) and \emph{dense} when they have many (Steam, MovieLens-1M).}
\label{tab:datasets}
\begin{adjustbox}{max width=\linewidth}
\begin{tabular}{llrrrrr}
\toprule
Role & Dataset & Users & Items & Interactions & per user & per item \\
\midrule
source & Amazon Video Games & 94,745 & 25,610 & 814,461 & 8.6 & 32 \\
source & Amazon Musical Instruments & 57,432 & 24,584 & 511,764 & 8.9 & 21 \\
source & Steam & 99,571 & 9,211 & 1,230,432 & 12.4 & 134 \\
held out & Amazon Industrial and Scientific & 50,985 & 25,847 & 412,938 & 8.1 & 16 \\
held out & MovieLens-1M & 6,040 & 3,416 & 999,611 & 165.5 & 293 \\
\bottomrule
\end{tabular}
\end{adjustbox}
\end{table}

\paragraph{Preprocessing.} Every dataset is reduced to its 5-core (i.e., we keep only users and items with at least 5 interactions each), with the interactions of each user ordered by time. The last item of each user is the test target, the one before it the validation target, and the rest is the training part, from which the baselines learn and the knowledge graph counts.

\subsection{Compared systems}
\label{sec:systems}

\paragraph{Variants of \model{}.} The three variants are named after the list that the decision model reranks, and each runs with a decision model of 0.8B or 4B parameters trained on the source domains. We write, for example, \fullsys{} (4B).
\begin{itemize}
\item \fullsys{}: all modules. SASRecF is one agent of the pool, the residual ranker keeps 30 candidates, and the decision model reranks them.
\item \listsys{}: the decision model reranks the 30 best items of SASRecF, with no pool. The difference from \fullsys{} is what the pool adds.
\item \kgsys{}: for a scenario with logs but no trained recommender, the decision model reranks the 30 candidates of the knowledge graph. In the held-out domains it also runs with a catalogue only and no logs.
\end{itemize}
Every variant gives two scores: the decision model \emph{read alone}, and the final score \emph{fused} with the order of the list it reranks.

\paragraph{Controls.} \emph{No decision model}: the list of each variant in its own order (for \fullsys{}, the residual ranker alone). \emph{Released decision model}: the same requests answered by the released Kev models, which, with behavior in the prompt and no training, is close to STAR \citep{lee2024star}. \emph{No statements}: \listsys{} with a decision model trained and read on item text only. \emph{Training in the held-out domain}: \listsys{} with a decision model trained on all four domains, as SASRecF is.

\paragraph{Baselines.} All baselines are trained and scored with one protocol over the whole catalogue (Table~\ref{tab:baselines}). The trained sequential models are GRU4Rec \citep{hidasi2016gru4rec}, SASRec \citep{kang2018sasrec}, BERT4Rec \citep{sun2019bert4rec} and SASRecF, a SASRec whose item representation adds embeddings of up to 12 catalogue attributes. All use the full softmax with cross-entropy. The training-free methods are popularity, item kNN on co-occurrence within five items, sequential rules \citep{ludewig2018evaluation}, content kNN on the text vectors, and a mix of transitions and text after \citet{han2026shortcut}. SASRecF is the strongest trained model on all users of the Amazon domains and Steam, so it is the trained recommender of the pool everywhere. It is not always the bar, however: on the development users of Steam SASRec is better, and on MovieLens-1M GRU4Rec. Every claim is therefore made against the \emph{best trained baseline of each domain}. The transfer baseline is PrepRec \citep{wang2024preprec}, whose items are represented only by their popularity over time. We train it with the authors' code on Video Games and Musical Instruments and score it on the held-out domains. Training details and the grid of each model are in the Additional Materials (AM4).

\begin{table}[t]
\centering
\caption{NDCG@10 of the baselines on all users of each domain (whole catalogue), best configuration of the full grid. Bold: best trained baseline. These values use all users, so they differ a little from the values on the 2,000 development users of Section~\ref{sec:results}. MovieLens-1M: the enriched version; with genres only, SASRecF reaches 0.1870 and is the best model.}
\label{tab:baselines}
\small
\begin{tabular}{lccccc}
\toprule
NDCG@10\hb{} & VG & MI & Steam & IND & ML-1M \\
\midrule
\multicolumn{6}{@{}l}{\emph{Training-free methods}} \\
popularity & 0.0126 & 0.0133 & 0.0259 & 0.0088 & 0.0180 \\
item kNN & 0.0516 & 0.0314 & 0.0475 & 0.0231 & 0.1337 \\
sequential rules & 0.0494 & 0.0287 & 0.0461 & 0.0226 & 0.1425 \\
content kNN & 0.0155 & 0.0111 & 0.0159 & 0.0150 & 0.0277 \\
transitions and text & 0.0250 & 0.0142 & 0.0274 & 0.0170 & 0.0370 \\
\addlinespace
\multicolumn{6}{@{}l}{\emph{Trained in each domain}} \\
GRU4Rec & 0.0538 & 0.0318 & 0.0637 & 0.0217 & \textbf{0.1853} \\
SASRec & 0.0551 & 0.0338 & 0.0672 & 0.0262 & 0.1850 \\
BERT4Rec & 0.0500 & 0.0298 & 0.0604 & 0.0200 & 0.1623 \\
SASRecF & \textbf{0.0585} & \textbf{0.0365} & \textbf{0.0680} & \textbf{0.0296} & 0.1828 \\
\bottomrule
\end{tabular}
\end{table}

\subsection{Protocol and metrics}

\paragraph{Metric.} Every method is scored by the rank of the held-out item over the whole catalogue, without the items of the history and without sampled negatives, which can change the order of methods \citep{krichene2020sampled}. A decision model ranks the 30 candidates it receives, and a target outside them counts as a miss, so its score includes the recall of its candidates. The main measure is NDCG@10: with one target at rank $r$ it is $1/\log_2(1+r)$ if $r \le 10$ and 0 otherwise. HR@10 and MRR@30 of the main systems are in the Additional Materials (AM10). In tables and figures, $\uparrow$ marks a measure for which higher is better and $\downarrow$ one for which lower is better.

\paragraph{Development users.} All design choices were made on development users: 2,000 test and 1,000 validation users per domain, drawn once with a fixed seed. Fusion weights are fitted on the validation users. %Every result of this paper is a development result. A test set that guides many choices slowly becomes optimistic, and reported gains in recommender systems have often failed independent checks \citep{ferraridacrema2019progress}.
We therefore reserved 10,000 fresh users per domain that no development step has used, with an analysis plan written before scoring them. T%his confirmation is part of the ongoing work (Section~\ref{sec:ongoing}). 
The lists of all users of every split will be released with their SHA-256 digests.

\paragraph{Significance.} We compare two systems on the same users with a paired randomization test, which \citet{smucker2007comparison} recommend for rankings. For each user $u$ we take the difference $d_u$ of NDCG@10 and change the sign of each $d_u$ at random, 10,000 times. The $p$-value is the share of these versions whose mean difference is at least as large, in absolute value, as the observed one. The test does not assume normal differences, which matters because NDCG@10 is zero for most users. Intervals are 95\% bootstrap intervals over users (2,000 resamples). In the tables, $^\ast$ marks $p < 0.05$ without correction. When we say ``not significantly different'' we give the interval, because a non-significant difference does not show that two systems are equal. The confirmation will control the family-wise error over nine hypotheses with the method of \citet{holm1979simple}.

\subsection{Implementation and release}

The decision models are Kev-0.8B, Kev-4B, Kev-9B and Kev-27B \citep{kev2026}, trained with Kev's trainer (adapters of rank 16; learning rate $10^{-4}$ for 0.8B, $5 \cdot 10^{-5}$ for 4B and 9B, and $2 \cdot 10^{-5}$ for 27B) and served through the System One protocol. The 0.8B and 4B models run on one GPU with 24 GB; the 9B and 27B models on one NVIDIA DGX Spark with 128 GB of unified memory. All answers of the decision models are stored with each run, and every number in this paper is computed from them. A cache of the requests also lets a run be repeated without the models. While preparing the release we found that the knowledge-graph agent broke ties in an order that depended on the hash seed of the Python process: between two processes this changed the candidates or the ``found by'' lines of 0.15 to 0.25\% of the test users and the NDCG@10 of the pool by at most 0.4\%. The released code fixes the order, and a test checks it.

The code (\url{https://github.com/marcopoli/decision4rec}) contains the framework, one backend for any model with the System One interface, the exporter of training records with the tests of its three rules, the configuration of every run, and guides to add a dataset, a decision model or an agent. The trained decision models are at \url{https://huggingface.co/m-polignano}. The datasets and the training records are rebuilt by scripts from their public sources. The answers of Jev are not redistributed, as its terms require. The folder \texttt{additional\_materials} of the repository holds the tables and analyses that do not fit here, cited as AM1 to AM11. AM11 lists the test behind every explanation given in this paper, with its script.

\section{Results}
\label{sec:results}

All numbers are on 5-core users (Section~\ref{sec:setup}), and every comparison is with the best trained baseline of the domain: SASRecF on the Amazon domains, SASRec on Steam and GRU4Rec on MovieLens-1M. NDCG@10 values are small because each test user has one correct item among 9,000 to 26,000, so we give most changes as percentages: $+19\%$ means 1.19 times the NDCG@10 of the reference. With 2,000 users per domain a difference below about 10\% is often not significant. Item kNN, for example, is 10\% below SASRecF on Video Games with p = 0.13.

\subsection{RQ1: domains whose records trained the decision model}

Table~\ref{tab:rq1} compares \model{} with the baselines of the three source domains. The rows of \model{} give the final score, the decision fused with the score of its candidates (Eq.~\ref{eq:fuse}).

\begin{table}[t]
\centering
\small
\caption{RQ1. The main result in the three source domains, 2,000 development test users per domain. NDCG@10, and change against the best trained baseline of the domain (first row). $^\ast$: significant difference (p $<$ 0.05). The rows of \model{} give the final, fused score. The 9B and 27B decision models were trained on a random 8,000 of the 34,667 records of the 0.8B and 4B models. \kgsys{} uses the 0.8B and 4B decision models only.}
\label{tab:rq1}
\begin{adjustbox}{max width=\linewidth}
\begin{tabular}{@{}lrrrrrr@{}}
\toprule
 & \multicolumn{2}{c}{Video Games} & \multicolumn{2}{c}{Musical Instr.} & \multicolumn{2}{c}{Steam} \\
\cmidrule(lr){2-3}\cmidrule(lr){4-5}\cmidrule(l){6-7}
 & NDCG@10\hb{} & change\hb{} & NDCG@10\hb{} & change\hb{} & NDCG@10\hb{} & change\hb{} \\
\midrule
Best trained baseline (SASRecF; on Steam SASRec) & 0.0622 & & 0.0370 & & 0.0790 & \\
Best training-free method (item kNN) & 0.0561 & $-10\%$ & 0.0337 & $-9\%$ & 0.0521 & $-34\%^\ast$ \\
\addlinespace
\multicolumn{7}{@{}l}{\emph{No trained recommender in the scenario}} \\
\kgsys{}, 0.8B decision model & 0.0637 & $+2\%$ & 0.0401 & $+8\%$ & 0.0558 & $-29\%^\ast$ \\
\kgsys{}, 4B decision model & 0.0630 & $+1\%$ & 0.0427 & $+15\%$ & 0.0668 & $-15\%^\ast$ \\
\addlinespace
\multicolumn{7}{@{}l}{\emph{The trained recommender is one agent of the pool}} \\
\fullsys{} without decision model (residual ranker only) & 0.0730 & $+17\%^\ast$ & 0.0468 & $+26\%^\ast$ & 0.0786 & $-1\%$ \\
\fullsys{}, 0.8B decision model & 0.0741 & $+19\%^\ast$ & 0.0467 & $+26\%^\ast$ & 0.0802 & $+2\%$ \\
\fullsys{}, 4B decision model & 0.0780 & $+25\%^\ast$ & \textbf{0.0504} & $+36\%^\ast$ & 0.0801 & $+1\%$ \\
\fullsys{}, 9B decision model (8,000 records) & 0.0736 & $+18\%^\ast$ & 0.0485 & $+31\%^\ast$ & 0.0798 & $+1\%$ \\
\fullsys{}, 27B decision model (8,000 records) & \textbf{0.0789} & $+27\%^\ast$ & 0.0492 & $+33\%^\ast$ & \textbf{0.0831} & $+5\%$ \\
\bottomrule
\end{tabular}
\end{adjustbox}
\end{table}

\paragraph{\fullsys{} is clearly better on the two Amazon domains.} With the Kev-4B decision model it is 25\% above SASRecF on Video Games and 36\% on Musical Instruments, and with the Kev-0.8B model 19\% and 26\% (all p $<$ 0.001). On Steam it is 1 to 2\% above SASRec, with a 95\% interval of $-6$ to $+10\%$, so we cannot tell it from the best model there. The gain thus appears in the sparse catalogues (16 to 32 interactions per item) and not in the dense one (134).

\paragraph{Most of the gain comes from the pool.} Without any decision model, the pool and its residual ranker are already 17\% and 26\% above SASRecF (p $\le$ 0.001). On top of the pool, the Kev-0.8B decision model adds 0 to 2\%, and the Kev-4B model adds 7\% on Video Games and 8\% on Musical Instruments (p = 0.026 and 0.013; 95\% intervals $+0.0006$ to $+0.0093$ and $+0.0007$ to $+0.0065$ NDCG@10). Read alone, the 4B model is 5\% and 13\% better than the 0.8B model (p = 0.030 and p $<$ 0.001). Over all six settings of this study, the 4B model adds more to the pool than the 0.8B model in five, and both add 2\% on Steam (Figure~\ref{fig:size}). Kev-27B, trained on a random quarter of the records (8,000 of 34,667), adds 4 to 8\% to the pool in all four domains, significantly on Video Games and Steam (p = 0.003 and 0.004), and both gains survive the Holm correction. On Steam, \fullsys{} (27B) is still not significantly different from SASRec ($+5\%$; 95\% interval $-2$ to $+14\%$).

\textit{Is the gain at 27B an effect of size or of data? }We trained the 0.8B and 4B models again on the same 8,000 records (Table~\ref{tab:datamatched}). On these records the gain over the pool grows with size on Video Games, Musical Instruments and Steam: 0 to 2\% at 0.8B, 2 to 4\% at 4B and 5 to 8\% at 27B. The 27B model is above the 0.8B model by 5 to 6\% in these three domains (p = 0.004, 0.049 and 0.006; Video Games and Steam survive the Holm correction) and by 1\% on Industrial and Scientific (p = 0.56). Against the 4B model it is significantly better only on Video Games ($+5\%$, p = 0.028). The 9B model does not continue the trend step by step: it adds 1 to 5\% to the pool, significantly in no domain (p $\ge$ 0.057), and it is within 2\% of the 4B model after fusion. The 27B model is above it on Video Games ($+7\%$, p = 0.001) and Steam ($+4\%$, p = 0.016). The gain grows with size on average, not at every step. More records help the 4B model but not the 0.8B model. Trained on all 34,667 records instead of 8,000, the 0.8B model changes by at most 1\% after fusion (p $\ge$ 0.39), while the 4B model gains 6 and 8\% read alone on Video Games and Musical Instruments (p = 0.007 and 0.049), and 4 and 5\% after fusion (p = 0.14 and 0.06). \textit{\textbf{Size therefore matters on the same data, and a larger model also makes better use of more records.}}

\begin{table}[t]
\centering
\small
\caption{Size and data. Fused NDCG@10 of \fullsys{} and, in brackets, the change against the same pool ordered by its residual ranker alone, 2,000 development test users per domain. The first four decision models were trained on the same random 8,000 records, the last two on all 34,667. $^\ast$: significant difference from the pool (p $<$ 0.05).}
\label{tab:datamatched}
\begin{adjustbox}{max width=\linewidth}
\begin{tabular}{@{}lllll@{}}
\toprule
NDCG@10\hb{} & Video Games & Musical Instr. & Steam & Industrial and Sci.$^\dagger$ \\
\midrule
pool, residual ranker only (reference) & 0.0730 & 0.0468 & 0.0786 & 0.0386 \\
\addlinespace
\multicolumn{5}{@{}l}{\emph{Trained on the same 8,000 records}} \\
Kev-0.8B & 0.0741\chg{$+2\%$} & 0.0467\chg{$0\%$} & 0.0794\chg{$+1\%$} & 0.0397\chg{$+3\%$} \\
Kev-4B & 0.0751\chg{$+3\%^\ast$} & 0.0479\chg{$+2\%$} & 0.0816\chg{$+4\%^\ast$} & \textbf{0.0406}\chg{$+5\%^\ast$} \\
Kev-9B & 0.0736\chg{$+1\%$} & 0.0485\chg{$+4\%$} & 0.0798\chg{$+2\%$} & \textbf{0.0406}\chg{$+5\%$} \\
Kev-27B & \textbf{0.0789}\chg{$+8\%^\ast$} & 0.0492\chg{$+5\%$} & \textbf{0.0831}\chg{$+6\%^\ast$} & 0.0402\chg{$+4\%$} \\
\addlinespace
\multicolumn{5}{@{}l}{\emph{Trained on all 34,667 records}} \\
Kev-0.8B & 0.0741\chg{$+2\%$} & 0.0467\chg{$0\%$} & 0.0802\chg{$+2\%$} & 0.0401\chg{$+4\%$} \\
Kev-4B & 0.0780\chg{$+7\%^\ast$} & \textbf{0.0504}\chg{$+8\%^\ast$} & 0.0801\chg{$+2\%$} & \textbf{0.0407}\chg{$+5\%$} \\
\bottomrule
\end{tabular}
\end{adjustbox}
\end{table}

\textit{How much of this can the training run explain?} We trained the 0.8B model twice more with other seeds of its trainer. The fused NDCG@10 of \fullsys{} moved by up to 3\% and the score read alone by up to 7\%, and 3 of the 24 paired tests between seeds were significant. A committee of the three seeds was not better than the best of them ($-2.6$ to $-0.9\%$, p $\ge$ 0.15). The step from 0.8B to 4B (5 and 8 points of gain over the pool on Video Games and Musical Instruments) is larger than this spread, while on Steam and Industrial and Scientific (0 and 1 point) it is not. The seed spread of the 4B model was not measured. \textit{The pool, which needs no language model at run time, is therefore the cheapest useful step, and the decision model is the part that improved with size.}

\paragraph{Without a trained recommender, \kgsys{} comes close to SASRecF.} On the two Amazon domains it is $+2\%$ and $+8\%$, not significantly different from SASRecF (95\% intervals $-10$ to $+16\%$ and $-8$ to $+27\%$), and significantly above item kNN, the best training-free method ($+14\%$ and $+19\%$, p $\le$ 0.014). On Steam it is 29\% below SASRec. Most of this level comes from the candidates of the knowledge graph: in their own order, with no model, they are at 99\% and 101\% of SASRecF (RQ3). A sparse scenario with logs but no training pipeline can thus come close to a tuned SASRecF without training anything, although with 2,000 users a loss of about 10\% cannot be excluded. With the 4B decision model, \kgsys{} is $+1\%$ and $+15\%$ on the two Amazon domains (95\% intervals $-11$ to $+15\%$ and $-1$ to $+34\%$), again not significantly different from SASRecF. On Steam the candidates already lose 31\% against SASRec. The 4B model recovers about half of this loss by ordering them better ($-15\%$, p = 0.009, against $-29\%$ at 0.8B), but no reranker can bring back a target that its candidates miss.

\subsection{RQ2: domains never used to train the decision model}
\label{sec:rq2}

On Industrial and Scientific we compare three situations: a catalogue only, logs but no trained model, and logs with a trained recommender. In the third, SASRecF and the residual ranker are trained in the domain, as they would be in any scenario, while the decision model is the one trained on the source domains.

\begin{table}[t]
\centering
\small
\caption{RQ2. Industrial and Scientific, a domain never used to train the decision model. NDCG@10, and its share of the NDCG@10 of SASRecF trained in this domain. $^\ast$: significant difference from SASRecF (p $<$ 0.05); not tested for PrepRec. The rows of \model{} give the final, fused score. $^\dagger$Best of the PrepRec models trained on Musical Instruments and on Video Games, with the item popularity of the test period, as in its paper; with only the popularity known at the time of the last item of the history, a stricter setting than for the other methods, which learn from the training interactions of all users, it reaches 0.0095 (31\%). The 9B and 27B decision models were trained on 8,000 records, and \kgsys{} uses the 0.8B and 4B decision models only.}
\label{tab:rq2}
\begin{adjustbox}{max width=\linewidth}
\begin{tabular}{@{}lrr@{}}
\toprule
System & NDCG@10\hb{} & share of SASRecF\hb{} \\
\midrule
\multicolumn{3}{@{}l}{\emph{A new scenario: catalogue only, no logs}} \\
content kNN & 0.0166 & $54\%^\ast$ \\
\kgsys{}, 0.8B decision model & 0.0211 & $68\%^\ast$ \\
\kgsys{}, 4B decision model & 0.0219 & $71\%^\ast$ \\
\addlinespace
\multicolumn{3}{@{}l}{\emph{Logs, but no trained recommender}} \\
PrepRec, a transfer model trained on another domain$^\dagger$ & 0.0123 & $40\%$ \\
item kNN, the best training-free method & 0.0262 & $85\%$ \\
\kgsys{}, 0.8B decision model & 0.0348 & $112\%$ \\
\kgsys{}, 4B decision model & 0.0353 & $114\%$ \\
\addlinespace
\multicolumn{3}{@{}l}{\emph{Logs, and a SASRecF trained in this domain}} \\
SASRecF & 0.0310 & $100\%$ \\
\fullsys{} without decision model (residual ranker only) & 0.0386 & $125\%^\ast$ \\
\fullsys{}, 0.8B decision model & 0.0401 & $129\%^\ast$ \\
\fullsys{}, 4B decision model & \textbf{0.0407} & $131\%^\ast$ \\
\fullsys{}, 9B decision model (8,000 records) & 0.0406 & $131\%^\ast$ \\
\fullsys{}, 27B decision model (8,000 records) & 0.0402 & $130\%^\ast$ \\
\bottomrule
\end{tabular}
\end{adjustbox}
\end{table}

\paragraph{A decision model trained elsewhere works here.} With its own SASRecF, \fullsys{} is 29 to 31\% above SASRecF (p $<$ 0.001), with a decision model that never saw the domain. The pool alone gives 25\%, and the decision model adds 4\% (0.8B) and 5\% (4B), not significantly on 2,000 users (p = 0.052 and 0.17). Training the decision model on this domain as well does not help. On the 30 best items of SASRecF (\listsys{}), a 0.8B model trained on all four domains is within 2\% of the one trained on three, here (p $\ge$ 0.57) and in the source domains (p $\ge$ 0.057). Only the cheap parts, SASRecF and the residual ranker, need training in a new scenario. The scales of the statements are not what makes this possible: a decision model that reads raw counts instead transfers as well (Section~\ref{sec:gap-scale}).

\paragraph{With logs only, the gain is not shown.} \kgsys{} is 12\% above SASRecF, with a 95\% interval of $-6$ to $+33\%$ (p = 0.20), and 33\% above item kNN (p $<$ 0.001). With the Kev-4B model it is 14\% above SASRecF ($-4$ to $+34\%$, p = 0.13). A loss larger than 6\% against a trained recommender is unlikely, but no gain is established: the logs, read by a model trained elsewhere, come close to a recommender trained here. PrepRec, built for transfer, reaches only 40\% of SASRecF, below the training-free methods that also use popularity. Its published gains are measured against 100 sampled negatives, while here every method ranks the whole catalogue. A model that reads only popularity over time therefore does not transfer to this domain. With a catalogue only, \kgsys{} is 27\% above content kNN (p = 0.002; 4B: 32\%, p $<$ 0.001) but reaches only 68\% of SASRecF (4B: 71\%): on its first day a scenario still needs logs.

\paragraph{MovieLens-1M.} The farthest domain from the sources has long histories and 293 interactions per item (Table~\ref{tab:ml1m}). On the plain version, with titles and genres only, \fullsys{} is not significantly different from GRU4Rec ($+1\%$ and $+3\%$; 95\% intervals $-4$ to $+6\%$ and $-1$ to $+9\%$). On the enriched version it is 5\% and 7\% above it (p = 0.042 and 0.007), the only significant gain in a dense domain. Most of this gain comes from the pool, which is 5\% above GRU4Rec by itself (p = 0.037). With the new attributes the agent of the knowledge graph finds the target for more users (7.3 against 5.6\%, p = 0.016), the text agent, which now reads the plots, for fewer (5.8 against 7.4\%, p = 0.008), and the pool as a whole for about as many (64.4 against 65.3\%, p = 0.14). The decision model changes the result of the pool by $-2$ to $+1\%$ (Figure~\ref{fig:size}), and the Kev-4B model is 2\% above the 0.8B model after fusion on both versions (p $\le$ 0.042). Trained also on 1,406 records of MovieLens-1M, the Kev-0.8B model repeats the ranker less, but read alone it becomes worse ($-3\%$ and $-4\%$, p = 0.07 and 0.006), and after fusion it stays at the level of the pool. Without a trained recommender, \kgsys{} read alone ranks 17 to 19\% below its own candidates at both sizes (p $<$ 0.001). With a catalogue only, the Kev-4B reader is 67 and 58\% above content kNN after fusion (p $<$ 0.001), against 36 and 41\% at 0.8B. The Kev-27B decision model, trained on the 8,000 records of Table~\ref{tab:datamatched}, puts \fullsys{} 5\% and 6\% above GRU4Rec (p = 0.055 and 0.010), but it adds only 1\% to the pool (p $\ge$ 0.07) and is not significantly different from the 4B model (p $\ge$ 0.17).

\begin{table}[t]
\centering
\small
\caption{MovieLens-1M, never used to train the decision model, 2,000 development test users. NDCG@10, and change against GRU4Rec, the best trained baseline on these users. Plain: titles and genres; enriched: also plots and Wikidata properties. $^\ast$: significant difference from GRU4Rec (p $<$ 0.05). The rows of \model{} give the final, fused score; the decision models are trained on the three source domains. $^\dagger$PrepRec trained on Video Games (better than Musical Instruments), with the popularity known at the time of the last item of the history (0.0207 with the popularity of the test period); it does not read item text, so both versions are the same; not tested.}
\label{tab:ml1m}
\begin{adjustbox}{max width=\linewidth}
\begin{tabular}{@{}lrrrr@{}}
\toprule
 & \multicolumn{2}{c}{Plain} & \multicolumn{2}{c}{Enriched} \\
\cmidrule(lr){2-3}\cmidrule(l){4-5}
System & NDCG@10\hb{} & change\hb{} & NDCG@10\hb{} & change\hb{} \\
\midrule
\multicolumn{5}{@{}l}{\emph{A new scenario: catalogue only, no logs}} \\
content kNN & 0.0282 & $-85\%^\ast$ & 0.0253 & $-86\%^\ast$ \\
\kgsys{}, 0.8B decision model & 0.0383 & $-79\%^\ast$ & 0.0357 & $-81\%^\ast$ \\
\kgsys{}, 4B decision model & 0.0472 & $-74\%^\ast$ & 0.0401 & $-78\%^\ast$ \\
\addlinespace
\multicolumn{5}{@{}l}{\emph{Logs, but no trained recommender}} \\
PrepRec, a transfer model trained on another domain$^\dagger$ & 0.0206 & $-89\%$ & 0.0206 & $-89\%$ \\
item kNN, the best training-free method & 0.1361 & $-26\%^\ast$ & 0.1361 & $-26\%^\ast$ \\
candidates of the knowledge graph, in their order & 0.1536 & $-17\%^\ast$ & 0.1499 & $-19\%^\ast$ \\
\kgsys{}, 0.8B decision model & 0.1532 & $-17\%^\ast$ & 0.1512 & $-18\%^\ast$ \\
\kgsys{}, 4B decision model & 0.1540 & $-16\%^\ast$ & 0.1500 & $-19\%^\ast$ \\
\addlinespace
\multicolumn{5}{@{}l}{\emph{Logs, and a SASRecF trained in this domain}} \\
GRU4Rec (best trained baseline) & 0.1844 & & 0.1844 & \\
SASRecF & 0.1806 & $-2\%$ & 0.1812 & $-2\%$ \\
\fullsys{} without decision model (residual ranker only) & 0.1910 & $+4\%$ & 0.1941 & $+5\%^\ast$ \\
\fullsys{}, 0.8B decision model & 0.1863 & $+1\%$ & 0.1939 & $+5\%^\ast$ \\
\fullsys{}, 4B decision model & 0.1907 & $+3\%$ & \textbf{0.1968} & $+7\%^\ast$ \\
\fullsys{}, 27B decision model (8,000 records) & \textbf{0.1938} & $+5\%$ & 0.1963 & $+6\%^\ast$ \\
\bottomrule
\end{tabular}
\end{adjustbox}
\end{table}

\subsection{RQ3: which part brings the gain}
\label{sec:rq3}

\paragraph{The pool.} The pool with its residual ranker brings most of the gain over the trained recommender: 17\% on Video Games, 26\% on Musical Instruments and 25\% on Industrial and Scientific (p $\le$ 0.001), and $-1\%$ on Steam (p = 0.89). The check of Section~\ref{sec:pool} shows that this gain comes from the features of the ranker, not from the candidates of the other agents. Starting from the score of SASRecF, the ranker was never more than 1\% below the best trained baseline in any of the six settings.

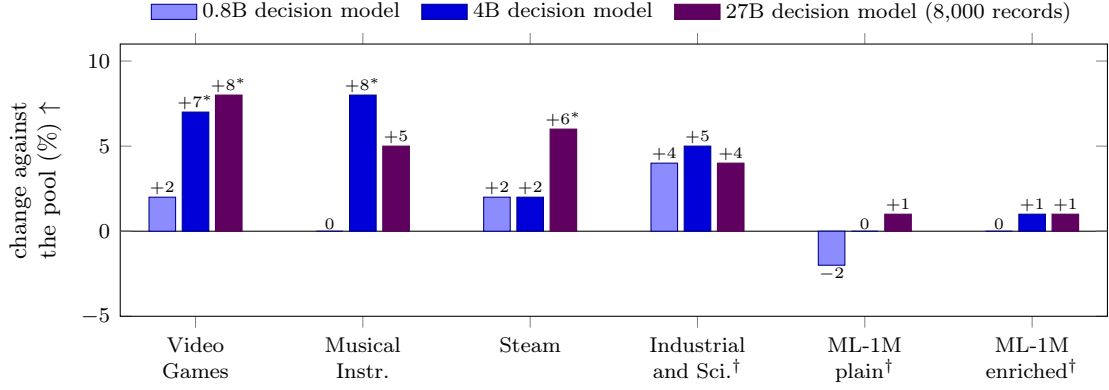
\begin{figure}[t]
\centering
\begin{tikzpicture}
\begin{axis}[ybar=2.5pt, scale only axis, width=0.82\linewidth, height=3.6cm, ymin=-5, ymax=11, bar width=10pt,
  ylabel={change against\\ the pool (\%) $\uparrow$}, ylabel style={font=\footnotesize, align=center},
  symbolic x coords={vg, mi, st, ind, mlp, mle}, xtick=data,
  xticklabels={Video\\ Games, Musical\\ Instr., Steam\\ \strut, Industrial\\ and Sci.$^\dagger$, ML-1M\\ plain$^\dagger$, ML-1M\\ enriched$^\dagger$},
  x tick label style={font=\scriptsize, align=center}, y tick label style={font=\scriptsize}, ytick={-5,0,5,10},
  point meta=explicit symbolic, nodes near coords, nodes near coords style={font=\tiny, inner sep=0.8pt}, area legend, enlarge x limits=0.09,
  visualization depends on={y \as \rawy}, every node near coord/.append style={anchor={\ifdim\rawy pt<0pt north\else south\fi}},
  legend style={font=\scriptsize, at={(0.5,1.04)}, anchor=south, legend columns=-1, draw=none, /tikz/every even column/.append style={column sep=6pt}},
  extra y ticks={0}, extra y tick labels={}, extra y tick style={grid=major, grid style={black}}]
\addplot[fill=blue!45!white, draw=dmline] coordinates {(vg,2)[$+2$] (mi,0)[0] (st,2)[$+2$] (ind,4)[$+4$] (mlp,-2)[$-2$] (mle,0)[0]};
\addplot[fill=blue!85!black, draw=dmline] coordinates {(vg,7)[$+7^\ast$] (mi,8)[$+8^\ast$] (st,2)[$+2$] (ind,5)[$+5$] (mlp,0)[0] (mle,1)[$+1$]};
\addplot[fill=violet!75!black, draw=rtline] coordinates {(vg,8)[$+8^\ast$] (mi,5)[$+5$] (st,6)[$+6^\ast$] (ind,4)[$+4$] (mlp,1)[$+1$] (mle,1)[$+1$]};
\legend{0.8B decision model, 4B decision model, {27B decision model (8,000 records)}}
\end{axis}
\end{tikzpicture}
\caption{What the trained decision model adds to the pool, by size. Each bar is the change of the fused NDCG@10 of \fullsys{} against the same pool ordered by its residual ranker alone (\fullsys{} without decision model), 2,000 development test users per domain. In five of six settings the 4B model adds more than the 0.8B model; on Steam both add 2\%. The 27B model, trained on 8,000 of the 34,667 records, adds 4 to 8\% in the four domains of Tables~\ref{tab:rq1} and~\ref{tab:rq2} and 1\% on both versions of MovieLens-1M. $^\ast$: significant (p $\le$ 0.026); the other differences are not significant on their own. Values from Tables~\ref{tab:rq3dm} and~\ref{tab:ml1m}.}
\label{fig:size}
\end{figure}

\paragraph{The decision model.} Table~\ref{tab:rq3dm} reads each trained decision model alone and after fusion, against the order of the list that it receives.
\begin{itemize}
\item \emph{By itself, it improves a trained recommender.} On the 30 best items of SASRecF, with no pool, the 0.8B model fused with SASRecF is 16 to 22\% above SASRecF in the three sparse domains (p $\le$ 0.012), and the 4B model 16 to 33\% (p $<$ 0.001). On Steam the Kev-4B model is 3\% above SASRec (p = 0.46). The Kev-4B model on the list of SASRecF reaches about the level of the pool (0.0730 on Video Games for both).
\item \emph{The statements bring most of the gain read alone.} Trained and read on item text only, a 0.8B decision model is 13 to 19\% below the one that also reads the statements on Video Games, Musical Instruments and Industrial and Scientific (p $\le$ 0.003), and 3\% below on Steam (p = 0.45). After fusion the gap is 4\% in three domains (p $\ge$ 0.14) and 16\% on Musical Instruments (p $<$ 0.001).
\item \emph{On top of the pool it adds more at 4B than at 0.8B} (Figure~\ref{fig:size}). The gain is not a correction that a stronger ranker would make. A residual ranker with one hidden layer improves the linear one only on Musical Instruments ($+4\%$, p = 0.013), and \fullsys{} (4B) stays 2.5 to 7.4\% above it. A LambdaMART ranker in LightGBM \citep{ke2017lightgbm}, on the same features, users and selection split, does not improve the linear one either ($-1.4$ to $+3.8\%$, p $\ge$ 0.25), and \fullsys{} (4B) is above it on Video Games, Musical Instruments and Industrial and Scientific ($+8.3$, $+9.3$ and $+6.7\%$; p = 0.016, 0.017 and 0.13) and not significantly different on Steam ($-1.8\%$).
\item \emph{On the candidates of the knowledge graph, too, the Kev-4B model adds more.} After fusion, the  Kev-0.8B model adds 3 to 13\%, significantly only on Industrial and Scientific, and the  Kev-4B model 3 to 23\%, significantly on Musical Instruments, Steam and Industrial and Scientific (p $\le$ 0.016, all three surviving the Holm correction over four domains). Head to head, the  Kev-4B model is 20\% above the  Kev-0.8B model on Steam (p $<$ 0.001) and within $-1$ to $+6\%$ of it elsewhere (p $\ge$ 0.075). The candidates themselves, in their own order, are not significantly different from SASRecF on the three sparse domains (99 to 101\%; 95\% intervals between $-17$ and $+20\%$).
\item \emph{The interface matters for training, not for accuracy.} A plain language model of the same size (\texttt{Qwen/Qwen3.5-0.8B-Base}), trained on the same records and read through the probabilities of option labels as in LlamaRec \citep{yue2023llamarec}, did not learn from 30 options in one pass: its scores stayed uniform. Read in blocks of 5 options and combined with the Luce model, it learned, and it was not significantly different from the Kev-0.8B decision model ($-3$ to $+3\%$ alone), with 12 questions per user instead of 2 and twice the time.
\end{itemize}

\begin{table}[t]
\centering
\small
\caption{RQ3. What the trained decision model adds to the list that it receives, 2,000 development test users per domain. Each block starts with the order of the list (reference); the other rows give NDCG@10 and, in brackets, the change against the reference. ``Alone'': the ranking of the decision model only. ``Fused'': fused with the score of the list, the final score. $^\ast$: significant difference from the reference (p $<$ 0.05). Bold: best value of each block.}
\label{tab:rq3dm}
\begin{adjustbox}{max width=\linewidth}
\begin{tabular}{@{}lllll@{}}
\toprule
NDCG@10\hb{} & Video Games & Musical Instr. & Steam & Industrial and Sci.$^\dagger$ \\
\midrule
\multicolumn{5}{@{}l}{\emph{\listsys{}: the decision model reranks the 30 best items of SASRecF (no pool)}} \\
order of SASRecF (reference) & 0.0622 & 0.0370 & 0.0759 & 0.0310 \\
Kev-0.8B, alone & 0.0655\chg{$+5\%$} & 0.0463\chg{$+25\%^\ast$} & 0.0687\chg{$-9\%$} & 0.0368\chg{$+19\%^\ast$} \\
Kev-0.8B, fused & 0.0722\chg{$+16\%^\ast$} & 0.0450\chg{$+22\%^\ast$} & \textbf{0.0821}\chg{$+8\%^\ast$} & 0.0370\chg{$+19\%^\ast$} \\
Kev-4B, alone & 0.0721\chg{$+16\%^\ast$} & 0.0486\chg{$+31\%^\ast$} & 0.0784\chg{$+3\%$} & \textbf{0.0372}\chg{$+20\%^\ast$} \\
Kev-4B, fused & \textbf{0.0730}\chg{$+17\%^\ast$} & \textbf{0.0491}\chg{$+33\%^\ast$} & 0.0815\chg{$+7\%$} & 0.0361\chg{$+16\%^\ast$} \\
\addlinespace
\multicolumn{5}{@{}l}{\emph{\fullsys{}: the decision model reranks the 30 best items of the pool}} \\
order of the residual ranker (reference) & 0.0730 & 0.0468 & 0.0786 & 0.0386 \\
Kev-0.8B, alone & 0.0752\chg{$+3\%$} & 0.0443\chg{$-5\%$} & 0.0804\chg{$+2\%$} & 0.0393\chg{$+2\%$} \\
Kev-4B, alone & \textbf{0.0790}\chg{$+8\%^\ast$} & 0.0499\chg{$+7\%$} & \textbf{0.0830}\chg{$+6\%$} & 0.0394\chg{$+2\%$} \\
Kev-0.8B, fused & 0.0741\chg{$+2\%$} & 0.0467\chg{$0\%$} & 0.0802\chg{$+2\%$} & 0.0401\chg{$+4\%$} \\
Kev-4B, fused & 0.0780\chg{$+7\%^\ast$} & \textbf{0.0504}\chg{$+8\%^\ast$} & 0.0801\chg{$+2\%$} & \textbf{0.0407}\chg{$+5\%$} \\
LlamaRec-style 0.8B, alone & 0.0727\chg{$0\%$} & 0.0454\chg{$-3\%$} & 0.0784\chg{$0\%$} & 0.0381\chg{$-2\%$} \\
LlamaRec-style 0.8B, fused & 0.0736\chg{$+1\%$} & 0.0459\chg{$-2\%$} & 0.0790\chg{$+1\%$} & 0.0396\chg{$+2\%$} \\
\addlinespace
\multicolumn{5}{@{}l}{\emph{\kgsys{}: the decision model reranks the 30 candidates of the knowledge graph (no trained recommender)}} \\
order of the candidates (reference) & 0.0614 & 0.0374 & 0.0542 & 0.0307 \\
Kev-0.8B, alone & 0.0590\chg{$-4\%$} & 0.0399\chg{$+7\%$} & 0.0598\chg{$+10\%$} & 0.0334\chg{$+9\%$} \\
Kev-0.8B, fused & \textbf{0.0637}\chg{$+4\%$} & 0.0401\chg{$+7\%$} & 0.0558\chg{$+3\%$} & 0.0348\chg{$+13\%^\ast$} \\
Kev-4B, alone & 0.0600\chg{$-2\%$} & 0.0419\chg{$+12\%$} & 0.0665\chg{$+23\%^\ast$} & 0.0344\chg{$+12\%$} \\
Kev-4B, fused & 0.0630\chg{$+3\%$} & \textbf{0.0427}\chg{$+14\%^\ast$} & \textbf{0.0668}\chg{$+23\%^\ast$} & \textbf{0.0353}\chg{$+15\%^\ast$} \\
\bottomrule
\end{tabular}
\end{adjustbox}
\end{table}

\paragraph{Why the Kev-0.8B model adds little to the pool.} Within the 30 candidates, the order of the 0.8B model has a Spearman correlation of 0.72 to 0.83 with the order of the residual ranker (Kev-4B: 0.66 to 0.77). On the list of SASRecF, where fusion adds 7 to 33\%, the correlation is only 0.32 to 0.37. Both orders are about equally good: the surprise of the next item, its negative log-likelihood among the 30 options, is 3.02 to 3.20 for the decision model, 3.07 to 3.24 for the ranker, and 3.40 for a uniform guess. The inputs of the ranker explain most of the order of the decision model. A gradient-boosted regression on the features of the ranker, fitted on half of the test users, explains 80 to 89\% of the variance of the scores of the 0.8B model within the list on the other half, and 67 to 82\% for the 4B model, less in every domain (p $<$ 0.001). Most of this agreement comes from the ``found by'' lines, which carry the ranks of the agents. Asked again without them (one question per user, 500 users per domain), the 0.8B model agrees less with the ranker (Spearman 0.67--0.71 falls to 0.22--0.42; 4B: 0.57--0.66 to 0.35--0.42; p $<$ 0.001 everywhere), while removing any one family of statements moves the agreement by at most 0.06. Two changes of training did not raise the fused score: leaving the ``found by'' lines out of a quarter of the records, and adding the score of the ranker to the logits in the loss, which removes the agreement. Giving the 0.8B model 250 characters per item instead of 90 changes its score read alone by $-2$ to $+7\%$ and after fusion by $-1$ to $+3\%$ (Additional Materials, AM6).

\paragraph{Which targets gain.} Table~\ref{tab:pop} splits the test users by how often their target was bought in the training data. In the three sparse domains \fullsys{} is above SASRecF for rare, middle and popular targets, significantly in eight of the nine groups. The largest relative gains are on rare targets in Video Games and Musical Instruments (from 0.0089 to 0.0235 and from 0.0037 to 0.0077), and on popular targets the gains are 9 to 31\%. On Steam no group changes significantly. Rare targets are among the 30 candidates of the pool for only 3 to 7\% of the users, against 31 to 44\% for popular ones, and NDCG@10 cannot exceed this share. In the sparse domains, then, the gain does not come at the expense of the long tail.

\begin{table}[t]
\centering
\small
\caption{Which targets gain: NDCG@10\hb{} on the 2,000 development test users of each domain, split by the purchases of the target item in the training data (rare: fewer than 20; middle: 20 to 99; popular: 100 or more). \fullsys{}: with the 4B decision model, fused. Users: share of the test users in each group. $^\ast$: significant difference from SASRecF within the group (paired randomization test, p $<$ 0.05). Bold: best system of each group.}
\label{tab:pop}
\begin{tabular}{@{}llrrr@{}}
\toprule
Domain & System & Rare & Middle & Popular \\
\midrule
Video Games & users & 30\% & 35\% & 35\% \\
 & SASRecF & 0.0089 & 0.0406 & 0.1292 \\
 & pool (no decision model) & $\mathbf{0.0240}^\ast$ & $0.0544^\ast$ & 0.1333 \\
 & \fullsys{} & $0.0235^\ast$ & $\mathbf{0.0615}^\ast$ & \textbf{0.1410} \\
\addlinespace
Musical Instr. & users & 44\% & 32\% & 24\% \\
 & SASRecF & 0.0037 & 0.0195 & 0.1234 \\
 & pool (no decision model) & 0.0056 & $0.0267^\ast$ & $0.1518^\ast$ \\
 & \fullsys{} & $\mathbf{0.0077}^\ast$ & $\mathbf{0.0295}^\ast$ & $\mathbf{0.1593}^\ast$ \\
\addlinespace
Steam & users & 7\% & 14\% & 79\% \\
 & SASRecF & 0.0168 & \textbf{0.0224} & 0.0909 \\
 & pool (no decision model) & 0.0233 & 0.0181 & 0.0945 \\
 & \fullsys{} & \textbf{0.0252} & 0.0183 & \textbf{0.0962} \\
\addlinespace
Industrial and Sci.$^\dagger$ & users & 59\% & 30\% & 11\% \\
 & SASRecF & 0.0127 & 0.0375 & 0.1111 \\
 & pool (no decision model) & 0.0145 & 0.0409 & $\mathbf{0.1612}^\ast$ \\
 & \fullsys{} & $\mathbf{0.0174}^\ast$ & $\mathbf{0.0482}^\ast$ & $0.1450^\ast$ \\
\bottomrule
\end{tabular}
\par\smallskip{\footnotesize }
\end{table}

\paragraph{Perspective agents.} We also tried agents that propose items from one point of view of the user (complement, continuation, replenishment, loyalty, exploration), in six versions (Additional Materials, AM5). Generated by a small language model, they add almost nothing. Built as rules over the knowledge graph, they find 5 to 10\% new targets and help the pool on Steam ($+4\%$, p = 0.016). With a decision model trained on pools that contain them, the whole system is 5\% better on Musical Instruments and 4\% on Steam than \fullsys{} (p = 0.021 for both), mostly through the pool. These results are exploratory, since several versions were tried on the same users.

\subsection{RQ4: trust and cost}

\begin{table}[t]
\centering
\small
\caption{RQ4. Can the probabilities of the trained Kev-0.8B decision model be trusted? Full architecture, 2,000 test users per domain. The last column gives the ideal value. Calibration error: expected calibration error (ECE) of the probability of the first choice. Surprise of the next item: negative log-likelihood (NLL) of the next item among the 30 options; a uniform guess gives 3.40. After scaling: after temperature scaling on validation users (fitted temperatures 0.95 to 1.20). Order change: mean deviation from a Luce model between the two orders of the same request. Position effect: the largest effect of a position on the log probability, against the first position. Conformal sets: for a target coverage of 90\%, on users whose next item is among the 30 options.}
\label{tab:trust}
\begin{tabular}{@{}lrrrrl@{}}
\toprule
 & Video G. & Musical I. & Steam & Industrial$^\dagger$ & ideal \\
\midrule
Calibration error, raw\lb{} & 0.019 & 0.015 & 0.015 & 0.031 & 0 \\
Calibration error, after scaling\lb{} & 0.030 & 0.025 & 0.015 & 0.031 & 0 \\
Surprise of the next item\lb{} & 3.08 & 3.17 & 3.20 & 3.07 & low; uniform: 3.40 \\
Order change\lb{} & 0.0015 & 0.0013 & 0.0010 & 0.0013 & 0 \\
Position effect & $-0.11$ & $-0.10$ & $-0.05$ & $-0.07$ & 0 \\
Conformal sets, coverage & 0.957 & 0.942 & 0.917 & 0.932 & 0.90 \\
Conformal sets, size (of 30)\lb{} & 24.8 & 25.3 & 24.2 & 23.0 & small \\
\bottomrule
\end{tabular}
\par\smallskip{\footnotesize }
\end{table}

\paragraph{The probabilities are calibrated but flat} (Table~\ref{tab:trust}). We measure them on the users whose target is among the 30 candidates, since a probability over the candidates cannot be right for the others. The calibration error is 0.015 to 0.031 without correction. Temperature scaling does not reduce it, because the temperature is fitted to the likelihood of the validation users and not to the calibration error. The surprise of the next item (3.07 to 3.20) is only a little below that of a uniform guess (3.40): a probability of the decision model can be read as a frequency, also in the held-out domain, but it rarely singles out one item. Conformal sets with 90\% coverage keep their promise (0.917 to 0.957) but need 23 to 25 of the 30 candidates, so they cannot shorten a list.

\paragraph{The order of the candidates has little effect.} The answers to the two orders of a request deviate from one Luce model by only 0.0010 to 0.0015, and correcting the largest position effect does not change NDCG@10 in practice, so answers to several requests can be combined.

\begin{table}[t]
\centering
\small
\caption{RQ4. The cost of one recommendation: one request per user with 30 candidates in two orders. Open models on one RTX 3090; Jev as measured through its API (Section~\ref{sec:jev}).}
\label{tab:cost}
\begin{tabular}{@{}llll@{}}
\toprule
 & Kev-0.8B, trained & Kev-4B, trained & Jev, paid API \\
\midrule
Time per user\lb{} & 157--167\,ms & 565\,ms & 310--340\,ms \\
Input tokens per request\lb{} & 3.9k--4.4k & 3.9k--4.4k & 3.6k--5.1k \\
Hardware & one 24\,GB GPU & one 24\,GB GPU & none \\
Money per 1,000 users\lb{} & GPU and electricity & GPU and electricity & \$0.15--0.22 \\
\bottomrule
\end{tabular}
\end{table}

\paragraph{One recommendation is one request.} The trained Kev-0.8B model needs 157 to 167\,ms per user on one consumer GPU, half the time of the hosted Jev and with no cost per request, and the 4B model 565\,ms (Table~\ref{tab:cost}). The decision model, not the pool, is the main cost, and \fullsys{} can rerank in real time.

\section{Jev Through Its Paid API, and Why We Train Kev}
\label{sec:jev}

Jev defined the interface that \model{} uses. Here we test what the commercial model gives inside \model{} and why the main experiments use Kev. We call the hosted \texttt{jev-1.13.0} (\url{https://typesafe.ai/}), pinned to one version, with exactly the requests of the Kev models: the same users, candidates, statements and ``found by'' lines. Jev cannot be trained by its users (the API offered no fine-tuning at the time of our runs, and its terms forbid training on its outputs), so it always runs as released. The fair comparisons are with the released Kev-4B and with the trained Kev-0.8B. All tests between two models are paired tests on the same users, and every answer of Jev is saved.

\begin{table}[t]
\centering
\small
\caption{The decision models of \model{} read alone, 2,000 development test users per domain. All models read the same requests. Each block starts with the order that the decision model receives (reference); the other rows give NDCG@10 and, in brackets, the change against the reference. Jev is used as released; Kev is used as released and trained on real choices of the three source domains, at 0.8B and 4B parameters. Bold: best model of the block. --: not applicable. The fused scores are in the Additional Materials (AM7). $^\dagger$Never used to train the decision models.}
\label{tab:jev}
\begin{adjustbox}{max width=\linewidth}
\begin{tabular}{@{}lllllll@{}}
\toprule
NDCG@10\hb{} & Video Games & Musical Instr. & Steam & Industrial$^\dagger$ & ML-1M plain$^\dagger$ & ML-1M enriched$^\dagger$ \\
\midrule
\multicolumn{7}{@{}l}{\emph{\fullsys{}: the decision model reranks the 30 best items of the pool}} \\
order of the pool (reference) & 0.0730 & 0.0468 & 0.0786 & 0.0386 & 0.1910 & 0.1941 \\
Jev, released & 0.0526\chg{$-28\%$} & 0.0356\chg{$-24\%$} & 0.0582\chg{$-26\%$} & 0.0344\chg{$-11\%$} & 0.1341\chg{$-30\%$} & 0.1372\chg{$-29\%$} \\
Kev-0.8B, released & 0.0565\chg{$-23\%$} & 0.0310\chg{$-34\%$} & 0.0555\chg{$-29\%$} & 0.0299\chg{$-23\%$} & 0.1409\chg{$-26\%$} & 0.1530\chg{$-21\%$} \\
Kev-4B, released & 0.0615\chg{$-16\%$} & 0.0348\chg{$-26\%$} & 0.0592\chg{$-25\%$} & 0.0334\chg{$-13\%$} & 0.1218\chg{$-36\%$} & 0.1332\chg{$-31\%$} \\
Kev-27B, released & 0.0668\chg{$-8\%$} & 0.0439\chg{$-6\%$} & 0.0636\chg{$-19\%$} & 0.0389\chg{$+1\%$} & 0.1521\chg{$-20\%$} & 0.1618\chg{$-17\%$} \\
Kev-0.8B, trained & 0.0752\chg{$+3\%$} & 0.0443\chg{$-5\%$} & 0.0804\chg{$+2\%$} & 0.0393\chg{$+2\%$} & 0.1844\chg{$-3\%$} & 0.1896\chg{$-2\%$} \\
Kev-4B, trained & \textbf{0.0790}\chg{$+8\%$} & \textbf{0.0499}\chg{$+7\%$} & \textbf{0.0830}\chg{$+6\%$} & \textbf{0.0394}\chg{$+2\%$} & \textbf{0.1859}\chg{$-3\%$} & \textbf{0.1902}\chg{$-2\%$} \\
\addlinespace
\multicolumn{7}{@{}l}{\emph{\kgsys{} with logs: the decision model reranks the 30 candidates of the knowledge graph}} \\
order of the candidates (reference) & 0.0614 & 0.0374 & 0.0542 & 0.0307 & 0.1536 & 0.1499 \\
Jev, released & 0.0465\chg{$-24\%$} & 0.0290\chg{$-22\%$} & 0.0475\chg{$-12\%$} & 0.0298\chg{$-3\%$} & 0.1006\chg{$-35\%$} & 0.0982\chg{$-34\%$} \\
Kev-0.8B, released & 0.0348\chg{$-43\%$} & 0.0185\chg{$-51\%$} & 0.0415\chg{$-23\%$} & 0.0221\chg{$-28\%$} & 0.0911\chg{$-41\%$} & 0.0856\chg{$-43\%$} \\
Kev-4B, released & 0.0466\chg{$-24\%$} & 0.0266\chg{$-29\%$} & 0.0486\chg{$-10\%$} & 0.0263\chg{$-14\%$} & 0.0953\chg{$-38\%$} & 0.0957\chg{$-36\%$} \\
Kev-0.8B, trained & 0.0590\chg{$-4\%$} & 0.0399\chg{$+7\%$} & 0.0598\chg{$+10\%$} & 0.0334\chg{$+9\%$} & 0.1248\chg{$-19\%$} & 0.1217\chg{$-19\%$} \\
Kev-4B, trained & \textbf{0.0600}\chg{$-2\%$} & \textbf{0.0419}\chg{$+12\%$} & \textbf{0.0665}\chg{$+23\%$} & \textbf{0.0344}\chg{$+12\%$} & \textbf{0.1258}\chg{$-18\%$} & \textbf{0.1249}\chg{$-17\%$} \\
\addlinespace
\multicolumn{7}{@{}l}{\emph{\kgsys{}, new scenario with a catalogue only: the reference is content kNN}} \\
order of the candidates (reference) & -- & -- & -- & 0.0166 & 0.0282 & 0.0253 \\
Jev, released & -- & -- & -- & 0.0186\chg{$+12\%$} & 0.0393\chg{$+39\%$} & 0.0354\chg{$+40\%$} \\
Kev-0.8B, released & -- & -- & -- & 0.0114\chg{$-31\%$} & 0.0305\chg{$+8\%$} & 0.0275\chg{$+9\%$} \\
Kev-4B, released & -- & -- & -- & 0.0130\chg{$-22\%$} & 0.0334\chg{$+18\%$} & 0.0298\chg{$+18\%$} \\
Kev-0.8B, trained & -- & -- & -- & 0.0205\chg{$+23\%$} & 0.0393\chg{$+39\%$} & 0.0361\chg{$+43\%$} \\
Kev-4B, trained & -- & -- & -- & \textbf{0.0223}\chg{$+34\%$} & \textbf{0.0475}\chg{$+68\%$} & \textbf{0.0391}\chg{$+55\%$} \\
\bottomrule
\end{tabular}
\end{adjustbox}
\end{table}

\paragraph{Training matters much more than size} (Table~\ref{tab:jev}). At the same size, the trained Kev-0.8B ranks 24 to 115\% better than the released Kev-0.8B in all fifteen settings (p $<$ 0.001). Five times more parameters without training (released Kev-4B against released Kev-0.8B) change the ranking by $-14$ to $+44\%$, significantly better in five settings and worse in two. At 4B, training raises the ranking on the candidates of the knowledge graph by 29 to 71\% (nine settings, p $<$ 0.001). A larger released model does not replace training either. Read alone on the pool, the released Kev-27B is 7 to 26\% better than the released Kev-4B (significantly in four of six domains) and 9 to 27\% better than Jev (in all six). The trained Kev-0.8B is still better than it in four of six domains (13 to 26\%, p $\le$ 0.005) and not significantly different in the other two ($+1\%$ in both), and training Kev-27B on 8,000 records raises its ranking by 19 to 32\% in four domains (p $<$ 0.001). The cheapest way to a useful decision model is to train a small one on real choices.

\paragraph{A small trained model beats the commercial one.} The trained Kev-0.8B is above the released Jev in eleven of fifteen settings, by 14 to 43\% (p $\le$ 0.042), and never below it. The other four settings are in the held-out domains: three are not significantly different ($+2$, $+11$ and $+12\%$) and one is equal. The trained Kev-4B is above Jev in fourteen of fifteen settings, by 15 to 50\% (p $\le$ 0.025), and not significantly different in the fifteenth ($+11\%$, p = 0.051).

\paragraph{As released, Jev is not significantly different from Kev-4B in ten of fifteen settings.} Kev-4B is ahead on Video Games with \fullsys{} (Jev 14\% lower, p = 0.003). Jev is ahead in four settings: the three scenarios with a catalogue only ($+43\%$, p = 0.002, Industrial and Scientific; $+18\%$, p = 0.027, and $+19\%$, p = 0.008, the two versions of MovieLens-1M) and the plain MovieLens-1M with \fullsys{} ($+10\%$, p = 0.006). These are the settings where the state carries no behavior of the scenario, or where the items are films whose titles language models are known to memorize \citep{dipalma2025memorize}.

\paragraph{Fusion makes weak models look alike.} After fusion, the three released models and the trained 0.8B model are within 0.0017 NDCG@10 of each other on the four main domains, against 0.0094 to 0.0249 when read alone, and all are close to the order of the pool. Only the trained 4B model stays clearly above, on Video Games and Musical Instruments. A comparison of decision models made only on fused scores would miss most differences.

\paragraph{What the API gives and what it does not.} The hosted model needs no GPU, no training data and no setup, so it is the fastest way to start \model{} in a new scenario. Only the input is paid (\$0.042 per million tokens at the time of our runs): 1,000 users cost \$0.15 to \$0.22, and this whole section cost \$4.89. An answer took 310 to 340\,ms. On the other side, the hosted model cannot learn the scenario, which Section~\ref{sec:whytrain} and RQ1 show to be the main need. Its outputs cannot train a cheaper model, the logs of the users leave the scenario with every request, and a version can be retired. Its answers are also not repeatable: we sent the same 1,200 requests again a few hours later, and only 0.3 to 0.7\% of the answers were identical, while the first choice changed in 7\% of the requests. The effect on accuracy is small, but a hosted result is not a function of the request alone (Additional Materials, AM7).

\paragraph{Why Kev.} Both models speak the same protocol, so replacing one with the other changes nothing in the architecture. Only the open model can be trained, and training is the step that improves the decision model most. We therefore keep Jev as the reference for a strong released decision model and use Kev for every trained component. A provider that allows training on the user's own data would remove the main limit of the hosted model.

\section{Discussion}
\label{sec:discussion}

\subsection{Limitations}

\paragraph{The decision model overlaps with the ranker.} On top of the pool, the trained Kev-0.8B decision model adds little: the inputs of the residual ranker explain 80 to 89\% of the variance of its order within the list, mostly through the ranks in the ``found by'' lines. At 4B the share is lower (67 to 82\%), and the model adds to the pool on two of four domains. A 27B model trained on a quarter of the records adds to it in all four domains, significantly on two, and on the same records it is above the 0.8B model in three domains. Its seed spread is not measured.

\paragraph{Dense domains.} On Steam and the plain MovieLens-1M, \fullsys{} is not significantly different from the best trained baseline, and \kgsys{} is below it. The only gain in a dense domain, 5 to 7\% on the enriched MovieLens-1M. Our results support \model{} for sparse catalogues, and for dense ones only with rich item content.

\paragraph{Without a trained recommender, the knowledge graph does most of the work.} In the sparse domains its candidates, with no model at all, cannot be told from SASRecF. On top of them the 0.8B decision model adds 3 to 13\%, significantly only on Industrial and Scientific, and the 4B model 3 to 23\%, significantly in three of four domains. Still, no version of \kgsys{} is significantly above the best trained baseline in any domain.

\paragraph{The ceiling of two stages.} Every reranker here orders 30 candidates, and the target is among them for about 9 to 25\% of the users. A better order raises accuracy inside this limit, not beyond it, and the flat probabilities of RQ4 show that the decision model rarely separates the target from the other candidates. Reading the 50 best candidates of the pool instead of 30, without new training, does not lift this limit either: after fusion the score changes by $-2$ to $+4\%$ at 0.8B and by $-1$ to $+4\%$ at 4B, significantly only once (Steam, 4B, $+4\%$, p = 0.045), which does not survive the Holm correction.

\paragraph{Evaluation protocol.} We hold out the last item of each user, as most work on sequential recommendation does. This split does not follow one global timeline, so every model, ours and the baselines, may learn from interactions that happened after a test item, which can change accuracy and even the order of models \citep{ji2023leakage}. All methods share the split, so the comparison is fair, but absolute values may differ under a split on global time. All baselines except PrepRec are our implementations, tuned on 4 to 9 configurations each, and models that transfer through text, such as RecGPT \citep{jiang2025recgpt}, are not included.

\paragraph{Few domains, few sizes, one run.} There are three source and two held-out domains, four sizes of decision model (0.8B, 4B, 9B and 27B), compared on the same 8,000 records, and one training run per model. For the 0.8B model, two more seeds moved the fused NDCG@10 by up to 3\%, so differences of this size between single runs are within the noise of training. Training on a fourth domain did not help \listsys{}. Whether many more scenarios would help is not tested.

\paragraph{Contamination and copying.} No dataset of this study can be assumed new for the language models we use, and no control can exclude every form of leakage. With the 0.8B decision model, on 2,000 users of all six settings, an empty history changes the score read alone by $-7$ to 0\% and a request without titles by $-6$ to 0\%, none of them significantly (p $\ge$ 0.05); after fusion the changes are $-5$ to $+1\%$. With the 4B model on MovieLens-1M, where memorized titles matter most, removing the titles costs 3\% after fusion on the plain version (p = 0.009) and nothing significant on the enriched one, and an empty history costs 2 to 3\% (p $\approx$ 0.05). Memory of the items therefore explains at most a few percent of our results. A model that reads a trained score can learn to copy it \citep{wang2026recallceiling}. Our decision model does not read the score of the trained recommender, but the ``found by'' lines give its rank, and they are the main source of the agreement with the ranker.

\paragraph{The hosted model.} Jev cannot be trained by its users, its answers are not repeatable, and its version can change, so its results can be checked only from the answers we saved. \model{} is not affiliated with TypeSafe, and no output of a hosted model was used to train any model.

\subsection{Ongoing work}
\label{sec:ongoing}

Several experiments are running or prepared, each tied to a claim above. (1) The confirmation of RQ1 and RQ2 on 10,000 fresh users per domain, with a split on global time and with SASRecF trained also on the targets that train the ranker and the decision model. (2) The data curve of the 0.8B model on 2,000 to 16,000 records, which has run and will be reported in the next version. (3) \fullsys{} with its decision model trained on all four domains. (4) More item text for the 4B model, a stronger text encoder in the pool, and a profile statement for long histories. (5) \fullsys{} inside the event-sourced runtime, with a check that replayed users receive the same rankings as in the batch pipeline. The status of each item is kept in the Additional Materials in the code repo.

\subsection{Take-home messages}

\begin{enumerate}
\item \emph{Train the decision model on real choices.} Released, the open Kev-4B helped a trained recommender no more than a random order did, a trained 0.8B model read the pool better than the released Kev-27B in four of six domains, and the released Jev was not significantly different from Kev-4B in ten of fifteen settings. A trained 0.8B model beats the released Jev in eleven of fifteen settings and is never worse.
\item \emph{Keep the trained recommender and correct its order.} A small residual ranker that uses what other agents know about each candidate gives most of the gain at almost no cost, and the candidates themselves can come from the trained recommender.
\item \emph{Measure the decision model alone, against the best baseline of each domain.} Fusion with the candidate score hides a weak model, and one baseline for all domains can hide a tie.
\item \emph{A larger trained model can add more.} At 4B the decision model added to the pool where the 0.8B model only matched it, on two of four domains, and a 27B model trained on a quarter of the records added to it in all four, significantly on two. On the candidates of the knowledge graph the 4B model added significantly in three domains, the 0.8B model in one. On the same 8,000 records the 27B model is still above the 0.8B model in three domains, so the gain is not only an effect of more data.
\item \emph{Expect gains in sparse catalogues.} In our domains the gain is large where items have few interactions and absent in a dense domain without rich item content.
\item \emph{Check the training records.} Records built from the candidates of another model can carry cues that exist only in training. The three rules of Section~\ref{sec:training}, each with a test, prevent this.
\end{enumerate}

\section{Conclusion}
\label{sec:conclusion}

Choosing the next item is a choice among options, and a decision model returns exactly one probability per option. This match is not enough on its own: as released, a decision model helps a trained recommender about as much as a random order of its candidates does. What it lacks is the behavior of the scenario, and \model{} adds it in three parts. A knowledge graph that is never trained writes the logs as statements on the scales of each scenario. A pool of agents keeps the list of the trained recommender and lets a small ranker correct its order. A decision model, trained once on real choices of three scenarios, reads candidates, text and statements through one interface and is reused unchanged in a fourth. In our experiments the pool gave most of the gain. The decision model added to it at 4B in two domains and at 27B in all four, significantly in two, and at 4B it also improved the candidates of the knowledge graph in three domains. On the same training records, the 27B model was still above the 0.8B model in three domains. Training mattered far more than the size or the price of the model. These results hold on development users of sparse catalogues. The confirmation on fresh users and the larger decision models are under way, with code, configurations and models open to anyone who wants to check or extend them.

\section*{Declarations}

\paragraph{Independence.} \model{} is an independent academic project. It is not affiliated with, endorsed by or connected to TypeSafe, the company that develops the Jev decision model, or to any of its products. Jev and the related names and trademarks belong to their owners. Jev is named in this paper only to describe the interface we study and the experiments we ran with it, through its paid public API. The framework is inspired by the interface of Jev and works with any decision model that has the same interface, such as the open Kev models. All trained models use open weights.

\paragraph{Data availability.} All datasets are public: Amazon Reviews 2023, the Steam dataset of \citet{kang2018sasrec}, MovieLens-1M, the plots of \citet{spillo2026bingewatch} and Wikidata. The scripts that rebuild every dataset, split and training record are in the repository. The user lists of all splits will follow, each with its SHA-256 digest.

\begin{sloppypar}
\paragraph{Code and models.} Code and configurations: \url{https://github.com/marcopoli/decision4rec}, with the Additional Materials in the folder \texttt{additional\_materials}. Trained models: \url{https://huggingface.co/m-polignano}.
\end{sloppypar}

\bibliographystyle{plainnat}
\bibliography{references}

\appendix
\section{Additional Details}
\label{sec:appendix}

\paragraph{Fitting the Luce model.} \label{app:luce} The parameters of Eq.~\ref{eq:luce} minimize $\sum_B \mathrm{KL}(p_B \,\|\, \pi_B)$. This is the cross-entropy of a multinomial logit with soft labels, convex in $(\theta, \beta)$, so it has no bad local solutions. Setting the derivative with respect to $\log v_i$ to zero gives $\sum_{B \ni i} p_B(i) = \sum_{B \ni i} v_i\, b_{\mathrm{pos}_B(i)} / Z_B$, and for fixed $Z_B$ the minorization-maximization update \citep{hunter2004mm}, with a weak Gamma prior of strength $a$, is
\begin{equation}
v_i \leftarrow \frac{a + \sum_{B \ni i} p_B(i)}{a + \sum_{B \ni i} b_{\mathrm{pos}_B(i)} / Z_B}, \qquad
Z_B = \textstyle\sum_{j} v_{i_j} b_j ,
\end{equation}
where $B \ni i$ runs over the questions that contain item $i$, $p_B(i)$ is the probability given to $i$ in $B$, and $\mathrm{pos}_B(i)$ is its position. The multipliers $\beta$ can be estimated only when an item appears in more than one position, which two orders guarantee. They belong to the decision model, not to the user, so they are estimated once on validation users. The mean residual $\frac{1}{|\mathcal{B}|} \sum_B \mathrm{KL}(p_B \,\|\, \pi_B)$ is zero if and only if the answers follow Eq.~\ref{eq:luce} exactly. Table~\ref{tab:trust} reports it as the order change.

\paragraph{Explanations.} \label{app:explain} For 500 test users per domain we asked the decision model of \fullsys{} again with one family of statements removed, or without the ``found by'' lines. The family whose removal changes the probability of the first item most is the \emph{explanation} of that recommendation. The explanations pass a deletion test: removing the chosen family changes the probability 3 to 4 times more than removing a family drawn at random (0.080 to 0.088 against 0.023 to 0.027 at 0.8B). But the explanation is the ``found by'' lines in 89 to 93\% of the cases, so these explanations say where a candidate came from. Follow-up, the strongest statement, explains 5\% of them. Without the ``found by'' lines the 0.8B model loses 15 to 33\% of its NDCG@10 read alone and the 4B model 3 to 22\%, while removing any single family of statements changes NDCG@10 by at most 6\%. A model trained to correct the ranker explains by behavior far more often (follow-up in 14 to 52\% of the cases) but does not rank better. The tables are in the Additional Materials (AM8).

\paragraph{Prompts.} The ranking instruction is the one used by \citet{lyu2026jevrerank}: ``Based on the user's interaction history, which candidate item is the user most likely to interact with next?''. After the history, the state has one line on the data of the scenario (``Store purchase data: full.'', ``Store purchase data: partial, from about 20\% of its customers.'' or ``Store purchase data: none, this store has a catalogue only.'') and a legend of the keywords with which the request writes the five families. In the order overall popularity, recent popularity, follow-up, co-purchase and text similarity, the legend reads: ``Candidate evidence, relative to this store: sold = all purchases; lately = last 30 days; after last = taken right after your last item; with yours = your items it is bought with; text = closeness to your last item.''. The prompts use the word ``store'' for a recommendation scenario, because the source domains are shops.

\end{document}